\documentclass[final,3p,times,twocolumn,authoryear]{elsarticle}

\usepackage{amssymb}
\usepackage{amsmath,amsfonts}
\usepackage{algorithmic}
\usepackage{algorithm}
\usepackage{array}
\usepackage[caption=false,font=normalsize,labelfont=sf,textfont=sf]{subfig}
\usepackage{textcomp}
\usepackage{stfloats}
\usepackage{natbib}
\setcitestyle{numbers,square}
\usepackage{url}
\usepackage{multirow}
\usepackage{CJKutf8}
\usepackage[table,xcdraw]{xcolor}
\usepackage{color}
\usepackage{graphicx}
\usepackage{tcolorbox}

\journal{Neurocomputing}

\begin{document}

\begin{frontmatter}



\title{ChatAgri: Exploring Potentials of ChatGPT on Cross-linguistic \\ Agricultural Text Classification}


\author{Biao Zhao\corref{cor1}\fnref{1}}
\ead{biaozhao@xjtu.edu.cn}

\author{Weiqiang Jin\corref{cor1}\fnref{1}}
\ead{weiqiangjin@stu.xjtu.edu.cn}
\cortext[cor1]{Both the first two authors, Biao Zhao and Weiqiang Jin, made equal contributions to this work.}


\author{Javier Del Ser\fnref{2}}
\ead{javier.delser@tecnalia.com}

\author{Guang Yang\corref{cor2}\fnref{3,4,5}}
\ead{g.yang@imperial.ac.uk}

\affiliation[1]{organization={School of Information and Communications Engineering, Xi`an Jiaotong University},
            addressline={Innovation Harbour}, 
            city={Xi`an},
            postcode={710049}, 
            state={Shaanxi},
            country={China}}



\affiliation[2]{organization={TECNALIA, Basque Research \& Technology Alliance (BRTA)},
            city={Derio},
            postcode={48160},
            country={Spain}}

\affiliation[3]{organization={Bioengineering, Imperial College London},
            city={London},
            postcode={SW7 2BX},
            country={UK}}

\affiliation[4]{organization={Imperial-X, Imperial College London},
            city={London},
            postcode={W12 7SL},
            country={UK}}

\affiliation[5]{organization={National Heart and Lung Institute, Imperial College London},
            city={London},
            postcode={SW3 6LY},
            country={UK}}

\cortext[cor2]{Corresponding author: Guang Yang.}  

\begin{abstract}
In the era of sustainable smart agriculture, a massive amount of agricultural news text is being posted on the Internet, in which massive agricultural knowledge has been accumulated. In this context, it is urgent to explore effective text classification techniques for users to access the required agricultural knowledge with high efficiency. 
Mainstream deep learning approaches employing fine-tuning strategies on pre-trained language models (PLMs), have demonstrated remarkable performance gains over the past few years. Nonetheless, these methods still face many drawbacks that are complex to solve, including: 1. Limited agricultural training data due to the expensive-cost and labour-intensive annotation; 2. Poor domain transferability, especially of cross-linguistic ability; 3. Complex and expensive large models deployment. 
Inspired by the extraordinary success brought by the recent ChatGPT (e.g. GPT-3.5, GPT-4), in this work, we systematically investigate and explore the capability and utilization of ChatGPT applying to the agricultural informatization field. Specifically, we have thoroughly explored various attempts to maximize the potentials of ChatGPT by considering various crucial factors, including prompt construction, answer parsing, and different ChatGPT variants. Furthermore, we conduct a preliminary comparative study on ChatGPT, PLMs-based fine-tuning methods, and PLMs-based prompt-tuning methods. A series of empirical results demonstrate that ChatGPT has effectively addressed the aforementioned research challenges and bottlenecks, which can be regarded as an ideal solution for agricultural text classification. Moreover, compared with existing PLM-based fine-tuning methods, ChatGPT achieves comparable performance even without fine-tuning on any agricultural data samples. We hope our preliminary study could prompt the emergence of a general-purposed  AI paradigm for agricultural text processing.
\end{abstract}



\begin{highlights}
\item Inspired by the success of ChatGPT, we propose ChatAgri, a ChatGPT-based approach for agricultural text classification.
\item We have designed several appropriate task-specific prompt inquiries strategies to intuitively trigger the understanding capability of ChatGPT based on ChatGPT prompt templates.
\item ChatAgri achieves competitive performance compared to existing PLM-based fine-tuning approaches, showing superior semantic understanding.
\item Zero-shot learning experiments demonstrate ChatAgri's potential for agricultural text classification, compared to existing PLM-based fine-tuning approaches.
\item Multi-linguistic experiments discussed demonstrate ChatAgri's excellent cross-linguistic transferability, enabling the model to adapt to different agricultural applications quickly.
\end{highlights}


\end{frontmatter}


\section{Introduction}
\label{intro}
With the rapid development of sustainable smart agriculture ecosystem, the quantity of various news contents related to agricultural themes on the Internet has undergone an explosive increase. Such a vast quality of unstructured data contains already latent historical knowledge, helping us precisely study natural hazards and mitigate potential agricultural risks. Artificial intelligence-based agricultural text classification enables managing these massive Internet agricultural news automatically and makes these massive unstructured data easily indexable, which is a crucial step for agricultural digitization and agricultural Internet of Things.

In recent years, these mainstream agricultural document processing techniques including text classification generally rely on various deep representation learning-based methods, especially on approaches based on pre-trained language models (PLMs), including BERT, BART, and T5 \cite{BERT,BART,T5}. Xu \textit{et al.} \cite{agriexporttextmin} proposed a novel model, namely time series-long short-term memory (AETS-LSTM), for predicting the rise and fall of agricultural exports. agricultural document processing. Cao \textit{et al.} \cite{BERT_agri_sentiana} utilized the BERT with symmetrical structure to analyze the sentiment tendency of the Internet consumers reviews towards the agricultural products. Leong \textit{et al.} \cite{agriOCR} employed a text-level character region awareness model (CRAFT) for recognizing and extracting the essential information from agricultural regulatory document and certificates. Jiang \textit{et al.} \cite{planthealthbulletin} proposed a BERT-based text classification network for automatically classifying the French bulletin to make these data easily indexable. In additional to the aforementioned research efforts, these deep representation learning-based approaches have held great promise for almost all agricultural informatization applications.  

Unfortunately, these PLMs-based fine-tuning solutions, inevitably encounter several challenging issues in the practical processes of model development, and application deployment. On the one hand, insufficient and poor quality supervised training data can greatly decrease the model performance whereas acquiring enough high-quality annotated data remains time-consuming and labour-intensive; on the other hand, even if trained properly on sufficient data, the inherent characteristics of supervised learning models limit their generalization capabilities to specific contexts related to the supervised corpus. In other words, when transplanted to new domains or new tasks, their limitations become evident, lacking a certain degree of scenario transferability, particularly of the cross-linguistic capacity. 
Moreover, due to the extremely large parameter volumes of PLMs, the corresponding deployment is complex and power-intensive, which requires high-performance equipments (such as massively parallel computing hardware, such as GPUs and TPUs). For example, the largest T5 model has over 11 billion parameters, which is 100 times the number of parameters of the BERT-base model. These prove that mainstream PLMs-based agricultural text classification methods fall far short of the standards for achieving General Purpose Artificial Intelligence (GPAI) in the future. 

\begin{figure}[!t]
\centering
\includegraphics[width=\linewidth]{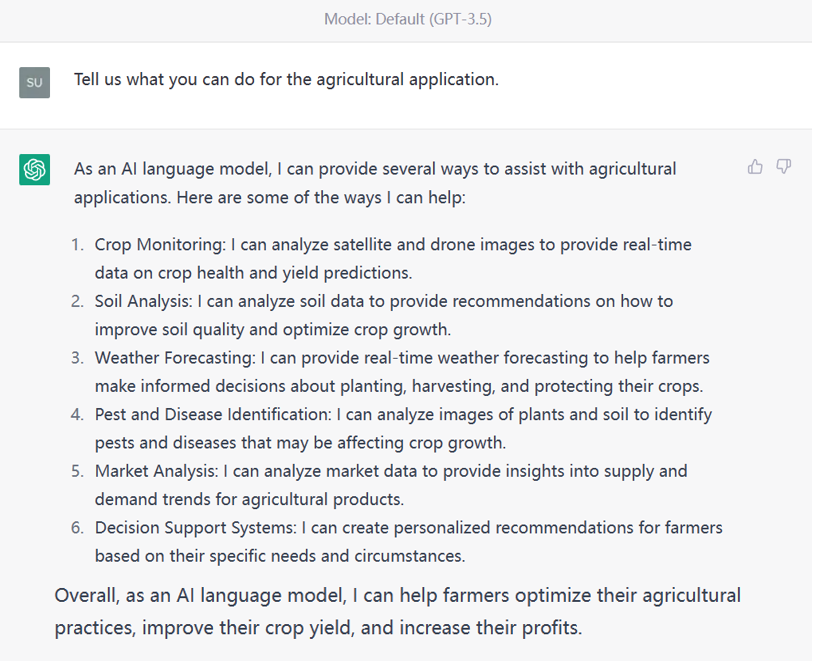}
\caption{Valuable suggestions advised by ChatGPT for assisting farmers and market regulator in better governing agricultural affairs (Query Date: 2023.3.16).}
\label{fig_ChatGPTcando}
\end{figure}

These introduced limitations and deficiencies have made existing agricultural document processing techniques can not handle well in almost application scenarios, especially for agricultural text classification.
Recently, the artificial intelligence ChatGPT-family chatbots, proposed by the OpenAI foundation, has caused a groundbreaking revolution in the academic community, especially for natural language processing (NLP) tasks. ChatGPT is essentially a powerful very large pre-trained language model for dialogue based on the Transformer architecture \cite{attentionneed}, utilizing a larger corpus, higher computational power, and an unprecedented amount of network parameters$\footnote{You can access ChatGPT by visiting the following URL: \\ \url{https://chat.openai.com/chat} [Accessed on 2023.05].}$. What is inspiring is that unlike previous intelligent chat robots, ChatGPT can provides smooth and comprehensive responses to various complex and professional human questions. For instance, ChatGPT can perform tasks such as multilingual translation, poetry generation, and code generation based on specific requirements \cite{eloundou2023gpts,GPT4}. Thus, ChatGPT have rapidly exhibited their remarkable language comprehension and generation abilities, which produces popularity and attracts ever-increasing attention in various cross-disciplinary researches that NLP community intersects with, such as radical radiology diagnosis \cite{radiologyreport_ChatGPT} and  sentiment analysis of surgery disease \cite{cvtassd,lymedisease_ChatGPT}.

After experiencing ChatGPT's universal and powerful capabilities, it is natural for us to wonder about how much potentials ChatGPT can bring to the agricultural products' production management process for optimizing sustainable agricultural applications. As shown in Fig.~\ref{fig_ChatGPTcando}, when asked about the potential applications of GPT-3.5 (a standard model in ChatGPT-family) in agriculture, The model replied that it is capable of performing tasks such as weather forecasting, pest and disease identification, and market analysis (among others).

\begin{figure*}[!t]
\centering
\includegraphics[width=\linewidth]{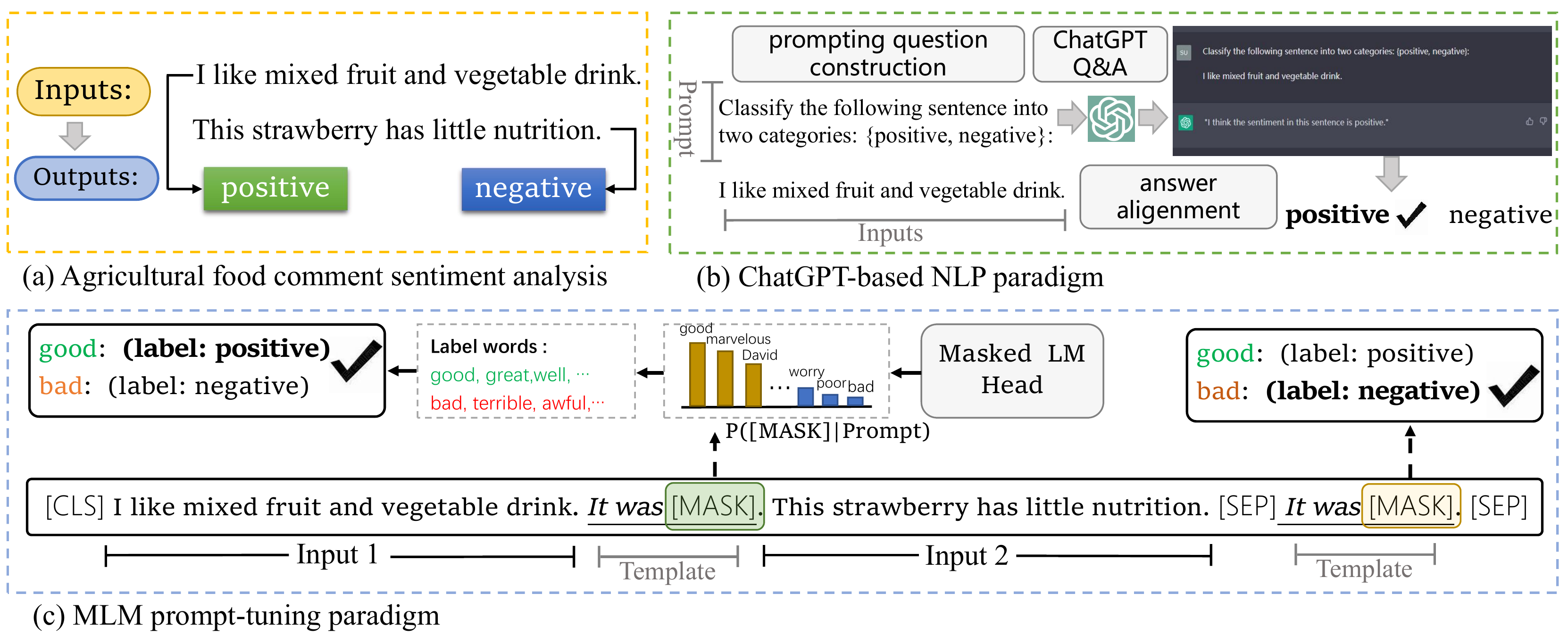}
\caption{The paradigm comparison of the ChatGPT-based NLP solutions and existing prompt learning paradigm using an agricultural sentiment analysis example. Part. (a) denotes the task prototype of the agricultural sentiment analysis; Part. (b) denotes the standard workflow of ChatGPT-based approaches; and Part. (c) denotes the standard workflow of Masked LM prompt-tuning methods.}
\label{fig_PromptDist}
\end{figure*}

Inspired by the potential applications of ChatGPT in the field of smart agriculture, it is our belief that the community is much in need for principled explorations to determine how much ChatGPT can contribute to the optimization of sustainable agricultural practices. With that concern in mind, we have decided to delve into the potentials of ChatGPT by focusing on the concise classification of agricultural text in this work.  

By doing so, our experiments mainly investigate the potential power of ChatGPT (i.e. GPT-3.5 by default) \cite{GPT3} and its extension (i.e. GPT-4) \cite{GPT4} for classifying the agricultural-related documents. Notably, along with the proposed ChatAgri, this paper also provides a brand-new paradigm which is distinguished from existing methods. Through a series comparative experiments of ChatAgri with a range of mainstream text classification models, including classic fine-tuned PLMs \cite{jwqrcekbqa,jwqknowABSA} and prompt-learning based on auto-regressive generative PLMs \cite{promptsurvey,sylNER,syl2022parallelnestedNER}, we systematically evaluated and investigated the superiority of ChatGPT in agricultural text classification tasks, which distinguished it significantly from other methods.

Furthermore, we have investigated extensive literature related to ChatGPT-based question answering (QA) \cite{twitter_ChatGPT,eventextr_ChatGPT,comparabert_ChatGPT,IE_ChatGPT} and the prompt learning scheme \cite{promptsurvey,ptuning,ptuningv2}, and arrived at the following conclusions: Most language understanding tasks based on ChatGPT can be categorized as a new form of Prompt Learning based on PLMs.
Specifically, regarding the adopted ChatGPT interface as a parameters-frozen large-scale PLM, the overall procedure are extremely similar to the prompt-tuning paradigm described in the survey of Liu \textit{et al.} \cite{promptsurvey}. Fig.~\ref{fig_PromptDist} gives a clear illustration of the major similarities and distinguishes between ChatGPT-based NLP paradigm, (a) and MLM prompt-tuning paradigm, (b), through a typical example of the agricultural food comment sentiment analysis task. As depicted in part. (c) of Fig.~\ref{fig_PromptDist}, the MLM prompt-tuning paradigm can be divided into three primary procedures: template engineering, pre-trained language models reasoning, and answer mapping engineering \cite{promptsurvey}. As shown in  part. (b) of Fig.~\ref{fig_PromptDist}, the general NLP research related to ChatGPT can be organized into the following several phases in our experiments \cite{radiologyreport_ChatGPT,survey_ChatGPT}: 1) prompting question construction engineering; 2) ChatGPT Q\&A inference; 3) answer normalization engineering (alias. answer alignment). Thus, several core factors were considered to be optimized: 

\begin{itemize}
\item 1). Due to that interacting with ChatGPT involves providing instructions through human response, based on previous ChatGPT prompting works \cite{NLGeva_ChatGPT,comparabert_ChatGPT,eventextr_ChatGPT}, we have designed several appropriate task-specific inquiries to intuitively trigger the understanding capability of ChatGPT; 

\item 2). As the textual generations of ChatGPT are essentially human-like natural language, they differ greatly when it comes to specific tasks. So, a accurate label mapping strategy from ChatGPT outputs to the final classified categories are needed to be developed. In our experiments, we devised two novel answer mapping strategies for this critical step for the answer alignment engineering.
\end{itemize}

To evaluate extensive data in various agricultural sub-fields, sourced mainly comes from Internet news covering topics ranging from insect pests, and natural hazards to agricultural market comments.
Further, even in cases multi-language corpora are tested, experiments validate that the proposed ChatAgri still features a significant transferring effectiveness in cross-linguistic scenarios.


In summary, our experiments provide a preliminary study of ChatGPT on agricultural text classification to gain a better understanding of it, and reported a systematic analysis according to the corresponding empirical results. We believe that by exploring how ChatGPT can contribute to agricultural production and management through text classification tasks such as pest and disease identification, agricultural news categorization, and market comment analysis, we can demonstrate the feasibility of ChatGPT in advancing agricultural practices, thereby paving the way for a more efficient and sustainable smart agriculture.

The novel ingredients of this work can be summarized as follows:
\begin{itemize}
\item Motivated by the various application progresses of very large pre-trained language models represented by ChatGPT, we conduct a preliminary study towards exploring the potentials of ChatGPT in agricultural text classification task and thus propose ChatGPT-based solution for agricultural text classification, namely ChatAgri;
\item Evaluated on several multi-linguistic datasets, ChatAgri achieves competitive performance compared to existing PLM-based fine-tuning approaches, showing a superior ability in terms of the impressive semantic understanding. Through several specific case analysis, it even surprisingly produces a intelligent reasoning chain;
\item The zero-shot learning experiments demonstrate the great potential of ChatAgri in agricultural text classification, compared to existing PLM-based fine-tuning approaches, which require high-quality supervised data, along with a time-consuming, labor-intensive annotations and expensive knowledge from agricultural domain experts;
\item Multi-linguistic experiments discussed in this work expose the excellent domain transferability of ChatAgri, by which the model can adapt to different agricultural applications quickly, and is a fundamental step accelerating the future General Purpose AI (GPAI); 
\item ChatAgri, only relying on network interface and minimum hardware requirements, subverts the mainstream complex and power-intensive PLM-based methods, which holds great promise of the general and low-costing artificial intelligence techniques for the future smart agricultural applications;
\item To encourage further research of smart agricultural applications by leveraging ChatGPT, we released the codes of ChatAgri on Github\footnote{Code has been released on Github: \url{https://github.com/albert-jin/agricultural_textual_classification_ChatGPT} [Accessed on 2023.05].}.
\end{itemize}

The remainder of this paper is organized as follows: Section~\ref{relatedwork} provides an overview of the recent literature in related fields, with a focus on recent research for the agricultural text classification task, ChatGPT, and pre-trained language model-based NLP techniques. 
Section~\ref{ChatAgri} presents a detailed description of the whole ChatAgri framework, including a detailed algorithmic description. In Section~ \ref{experimentssetup} and \ref{experiments}, we conduct a comprehensive analysis of the comparison experiments between ChatAgri and several mainstream PLM-based methods, along with various ablated studies. Finally, Section~\ref{conclusion} offers a concise summary of the primary contributions of our research and outlines future prospects for further sustainable smart agriculture development based on our findings. 


\section{Related Work}
\label{relatedwork}
In this section, we will review the related literature on accurately classifying cross-linguistic agricultural texts, recent advancements and applications in ChatGPT and its extensions, as well as PLM-based fine-tuning and prompt-tuning approaches in addressing the challenges of agricultural text classification.

\subsection{Agricultural Text Classification}
Over the past decade, the primary machine learning models (e.g. decision tree, CNN, LSTM, and GRU) \cite{agriexporttextmin} have been the dominant approaches in research on the agricultural document classification. 

Azeez \textit{et al.} \cite{agriieee} used the support vector machine (SVM) and decision tree induction classifiers to complete the regional agricultural land texture classification. Li \textit{et al.} \cite{ieeeaccdff} simultaneously utilized the Bi-LSTM and the attention mechanism to further dynamically enrich the extracted multi-sources semantic features, which effectively improve the performance of agricultural text classification. Dunnmon \textit{et al.} \cite{foodsecurity} leveraged CNN to predict agricultural Twitter feeds from farming communities to forecast food security indicators, and demonstrated that CNNs are widely superior to RNNs in agriculturally-relevant tweets sentiment classification.

Since the introduction of large models such as BERT \cite{BERT} and GPT \cite{GPT1}, many NLP tasks have achieved significant performance improvements and have gradually replaced traditional machine learning approaches \cite{survey_ChatGPT}. Compared to traditional machine learning methods, large pre-trained language models are better equipped to handle the complexity scenarios, having received widespread attentions in both academic and industrial settings. 

Shi \textit{et al.} \cite{shiBERTagri} employed BERT to identify the most representative information from unlabeled sources, which were manually labeled to construct the corpora of agricultural related news from diversified topics, enhancing the efficiency of labeling process and ultimately improving the corpora construction quality. Jiang \textit{et al.} \cite{planthealthbulletin} automatically classify the French plants health bulletins to make these data easily searchable through fine-tuning BERT. Leong \textit{et al.} \cite{agriOCR} developed an automatic optical character recognition system for the categorization and classification of agricultural regulatory documents. To tackle the imbalance between the supply and demand of the agricultural market, Cao \textit{et al.} \cite{BERT_agri_sentiana} introduced a improved BERT-based sentiment analysis model for agricultural product evaluation through Internet reviews. The proposed BERT model with symmetrical structure accurately identifies the emotional tendencies of consumers, helping consumers evaluate the quality of agricultural products and helping agricultural enterprises optimize and upgrade their products. 

\subsection{Traditional Machine Learning methods, and \\ PLM-based Fine-tuning, and Prompt-tuning}
For a significant period of time in the past, the predominant approach for addressing the agricultural text processing problems was based on traditional machine learning methodologies. Xu \textit{et al.} \cite{agriculturalexpo} proposed a novel method to predict the rise and fall of agricultural exports, called agricultural exports time series-long short-term memory (AETS-LSTM). AETS-LSTM achieves improved prediction performance that predicts the tendencies of the agricultural exports, which is effective way to help agribusiness operators to make better evaluations and adjustment policies. To identify the pests and diseases symptoms of rice farming,  Costa \textit{et al.} \cite{pestclassifi} build a knowledge-based system that used jaccard similarity coefficient (JSC), which performs tokenizing, filtering and porter stemming to extract critical information to deliver pests and disease problem.

Feature engineering-based methods were limited by their inability to capture the complexity and nuances of natural language, particularly when it comes to some semantic complex situations \cite{survey_ChatGPT}. With the emergence of PLMs \cite{GPT1,GPT2,BERT,BART}, a powerful technique that revolutionized the field of NLP, many traditional methods \cite{planthealthbulletin,jwqdasfaa,foodsecurity} has been substituted \cite{attentionneed}. Since then, the PLM-based fine-tuning paradigm has been propelled to be the mainstream learning technique for various agricultural information processing \cite{deeplearning_agri}. PLM-based fine-tuning paradigm is designed by introducing additional network parameters and fine-tuning PLMs to downstream tasks using task-specific objective functions.
Cao \textit{et al.} \cite{BERT_agri_sentiana} developed an improved BERT-based model to extract complete semantic information for the task of sentiment analysis in agricultural product reviews. The goal was to assist consumers in making informed purchasing decisions. They utilized TensorFlow to fine-tune the whole parameters of BERT and its downstream classifier to obtain a well-optimized model. Jin \textit{et al.} \cite{jwqknowABSA} proposed a dictionary knowledge infused network, DictABSA, for sentiment analysis and agricultural text classification.

Nevertheless, these PLM-based fine-tuned models may not generalize well to new scenarios and required significant amount of annotated data, making it hard to be quickily developed and easily deployed. As a result, the role of traditional PLM-based fine-tuning has gradually diminished in NLP, being replaced by a more promising learning paradigm known as ``prompt learning'' or ``prompt-tuning'', according to a recent survey \cite{promptsurvey}. Different from the PLM-based fine-tune paradigm, prompt-tuning follows the original LM training which adapts the downstream task to the PLM itself with the help of constructed prompting templates, thus especially performing well in few-shot or even zero-shot scenarios. Lyu \textit{et al.} \cite{radiologyreport_ChatGPT} investigate the effect of different optimized prompts on the performance of the improved plain-language translations of the radiology report. Liu \textit{et al.} \cite{ptuning} proposed \textit{P-Tuning}, a novel method that automatically searches for prompts in the continuous space to improve the performance of PLMs. It uses a few continuous free parameters as prompts and optimizes them using gradient descent. Experiments proved that \textit{P-Tuning} brings substantial improvements to GPTs, even outperforms BERT models to some extent. Liu \textit{et al.} \cite{ptuningv2} also introduced \textit{P-Tuning v2}, a enhanced continuous prompt optimization method of \textit{P-Tuning} \cite{ptuning}. \textit{P-Tuning v2} represents a significant improvement over \textit{P-Tuning} by using continuous prompts for every layer of the PLM, rather than just the input layer, increasing the capacity of continuous prompts and helping to close the gap to fine-tuning across the small models and hard tasks. Hu \textit{et al.} \cite{knowledgeableVerbalizer} devised a novel knowledge enhanced method for text classification, namely \textit{knowledgeable prompt-tuning} (KPT). It incorporates rich external knowledge from knowledge bases (KBs) into the prompt verbalizer to better stimulate the internal knowledge in PLMs.

\subsection{ChatGPT}
ChatGPT is a leading conversational language model developed by OpenAI, which serves as an expert in all fields with omnipotent and omniscient knowledge. ChatGPT is a disruptive revolution across numerous research domains, extending beyond NLP, providing a user-friendly interface that grants the general public unprecedented access to the capabilities of large language models. ChatGPT, also known as GPT-3.5 that built upon GPT-3 \cite{GPT3}, serves as a conversational robot capable of comprehending intricate instructions and producing high-quality replies across diverse scenarios. ChatGPT, acting as a valuable tool, has made a significant contribution to many application scenarios and has opened up new possibilities for virtual assistants.
In terms of model structure, ChatGPT \cite{GPT4,eloundou2023gpts} can be regarded as a quantum leap characterized by several distinctive characteristic features that stands out from previous NLP models such as BERT \cite{BERT}, BART \cite{BART}, and T5 \cite{T5}. These can be summarized as: a very large language model using over billions of parameters, having the capability of a chain of thought prompting, and trained with reinforcement learning from human feedback (RLHF). 

As millions of users continue to tap into these language models, countless new use cases emerge, opening the door to a flurry of ChatGPT potentials.
Based on a recent empirical study \cite{NLGeva_ChatGPT,survey_ChatGPT}, ChatGPT has shown remarkable proficiency in multilingual translations, particularly in high-resource languages translator such as mutual translation between various European and American languages. Furthermore, this study found that ChatGPT performs similarly to other prominent translation services like Tencent TranSmart, DeepL Translate, and Google Translate.
What's even more impressive is that ChatGPT can be used in code debugging and even code generation \cite{GPT4}. Based on Haque \textit{et al.} \cite{twitter_ChatGPT}, ChatGPT was evaluated on its capability to provide code snippets that adhered to the syntax and semantics of the programming language, such as Python, Java, and JavaScript. Bang and colleagues \cite{ChatGPT_code} utilized several codes, including Python Turtle graphics and HTML Canvas, acted as tools for the multimodal task of generating images from text. These researchers demonstrate that ChatGPT was able to generate superior-quality codes based on brief business requirements expressed in natural language, overwhelmingly surpassing other code modification techniques.

The growing fascination with ChatGPT has spurred a wide range of investigations into the myriad of possibilities presented by this groundbreaking language model, particularly those in the agricultural field.  
Gao \textit{et al.} \cite{eventextr_ChatGPT} investigates the feasibility of using ChatGPT for event extraction, highlights the difficulties posed by event extraction due to its complexity and the need of a comprehensive set of instructions. Wei \textit{et al.} \cite{IE_ChatGPT} designed a universal zero-shot information extraction framework via chatting with ChatGPT, namely ChatIE, handling NLP tasks including named entity recognition, event extraction, and relation extraction. 
Specifically, ChatIE is devised as a decomposed multi-stages involving with several turns of QA: first stage to discover the element types presented in the sentence through one turn of QA, and second stage to find the elements to fill the corresponding element slots through multiple QA turns.

Furthermore, OpenAI \cite{GPT4} released GPT-4, an advanced, large-scale, multi-modal generative PLM, in early March of this year. It exhibits significant improvements over ChatGPT (GPT-3.5) in terms of multi-modal image and text interaction, broader digital character limitations and more accurate semantic understanding. GPT-4 holds immense promise for future diverse applications and is regarded as a significant stride toward achieving general-purpose technology. 

Moreover, the official investigation of GPT-4 \cite{eloundou2023gpts} confirms the hypothesis that these technologies can have a substantial effect on a wide swath of occupations, especially for higher-wage occupations that face greater exposure to PLMs. Recently, an open letter signed by numerous prominent researchers has called for a halt to ``Pause Giant AI Experiments'' towards the successive development of GPT-5 due to GPT-4's perceived terrifying power and its potential risks to society \cite{GPT4}. Even Sam Altman, the CEO of OpenAI, has also signed this open letter, demonstrating that the future impact of General Purpose AI, represented by ChatGPT, on various industries will be revolutionary and profoundly impressive.

\section{ChatAgri: ChatGPT-based Agricultural Text Classification} 
\label{ChatAgri}

\begin{figure*}[!t]
\centering
\includegraphics[width=\linewidth]{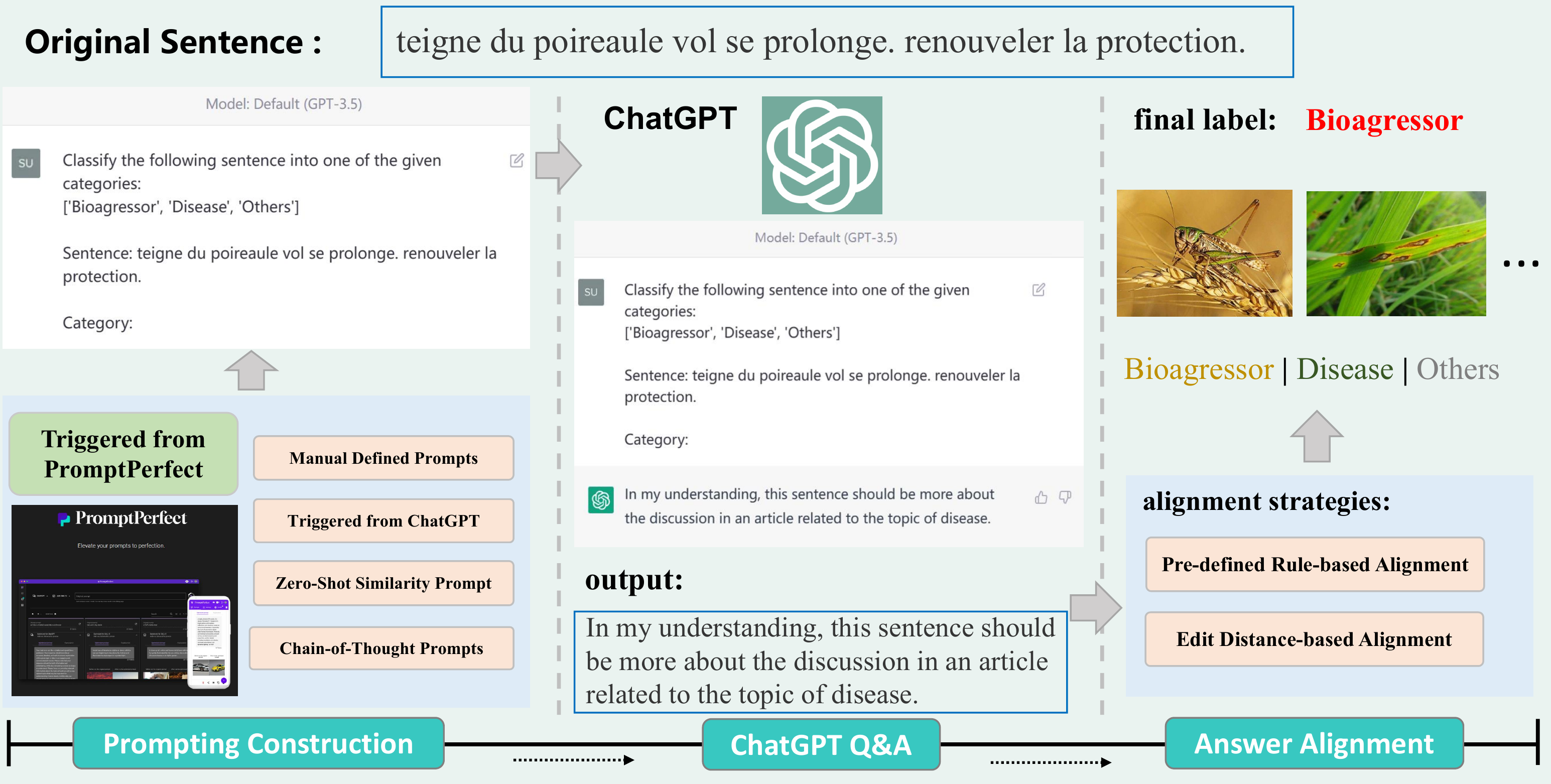} 
\caption{The framework of ChatAgri, which is illustrated by an typical example in the agricultural natural disaster dataset, French Plant Health Bulletin. First (left), several prompting construction strategies were applied to generate prompts, and the ChatGPT question is constituted by integrating these prompts with the original sentence; Second (center), ChatGPT provides response based on the inputs; Finally (right), the answer alignment strategies were devised to classify the intermediate answer to pre-defined categories.}
\label{fig_ChatAgri}
\end{figure*}

\subsection{Methodology Overview}
Focusing on investigate the feasibility of applying ChatGPT to agricultural text classification, ChatAgri, one of the preliminary studies of ChatGPT-based agricultural applications is constructed in this paper, along with a series of systematically and exploratory experimental analysis discussed.

Through our investigations, there are no existing research works that systematically utilized ChatGPT to the text classification task until our ChatAgri proposed. To fill this gap, the question how to defined the corresponding general workflow for the ChatGPT-based agricultural text classification will be further discussed. Specifically, after referred to abundant latest literature, as shown in Fig.~\ref{fig_ChatAgri}, we deem that almost all the ChatGPT-assisted applications can be divided into three phrases: 
\begin{itemize}
\item Prompting Question Construction: The first stage which focuses on providing appropriate prompting strategies to be fed into ChatGPT;
\item ChatGPT Q\&A Inference: The second stage about the reasoning procedure of ChatGPT Q\&A, which is transparent to us and can be regarded as a black box;
\item Answer Normalization or Alignment: The third stage transferring the natural language intermediate response to the target label in the pre-defined categories.
\end{itemize}

Among these steps, in additional to the Q\&A inference conducted by ChatGPT, a static reasoning procedure we can not participate in modification, the prompting construction engineering and answer alignment engineering can be further optimized during our experiments. From a macro perspective, ChatAgri is a pipeline structure in which each procedure influence the final prediction performance to a certain extent, including the quality of constructed prompts, the selected ChatGPT version, and the priority of adopted answer mapping strategies. Thus, the next subsections will introduce multiple novel solutions which are utilized in our experiments to fully exert the enormous potential and superiority of the ChatGPT in ChatAgri.  

Furthermore, as opposed to the text classification in the universal domain, the agricultural text classification acted as a domain-specific research branch due to the additional requirements of domain expertise knowledge. Another crucial factor, domain-specificity, should also taken into more considerations and corresponding customized strategies.

The following chapters would successively elaborate the specific solutions during the entire experiments of ChatAgri.  

\subsection{Prompt Question Construction}
\label{pqc_sub}
It is widely acknowledge to us that prompting engineering is a cumbersome art that requires extensive experience and manual trial-and-errors \cite{promptsurvey,survey_ChatGPT}. To design the suitable prompts to trigger the sentence classification ability of ChatGPT, we investigate sufficient pioneering works that discuss about how to generate optimized ChatGPT prompting questions \cite{trans_chatgpt,eventextr_ChatGPT,comparabert_ChatGPT}. Specifically, as depicted in the left of Fig.~\ref{fig_ChatAgri}, the adopted prompt generation strategies in this experiments includes: 1). manually defined prompts; 2). prompts triggered from ChatGPT; 3). prompts based on the zero-shot similarity comparisons; and 4). prompts based on Chain-of-Thought (CoT); 
These novel prompt generation strategies are discussed in the followings.

\subsubsection{Manually Defined Prompts}
\label{mdp_subsub}
Following the general communication habits, we manually elaborate several prompting templates, Table~\ref{table_manuallyprompts} displays the part of designed prompts. Note that it is necessary to provide ChatGPT with the two mentions: original textual context and pre-defined categories, through some appropriate ways. Furthermore, for simplicity, we insert two extra slots into the prompts to combine the corresponding mentions, which respectively are \texttt{[SENT]} (slot of sentence) and \texttt{[CATE]} (slot of categories). 

\begin{table}[h]
\renewcommand{\arraystretch}{1.3}
\caption{The partial manually devised prompts. \texttt{[Res]} denotes the response provided by ChatGPT. \label{table_manuallyprompts}}
\centering
\scalebox{0.75}{
\begin{tabular}{cc}
\hline
\textbf{No.} & \textbf{prompting template} \\ \hline
\textbf{1} &
  \begin{tabular}[c]{@{}c@{}}Classify the following sentence into one of the given \\ categories: \texttt{[CATE]} \textcolor[rgb]{1,0,0}{$\backslash$n} Sentence: \texttt{[SENT]} \textcolor[rgb]{1,0,0}{$\backslash$n} Category: \textcolor[rgb]{1,0,0}{$\backslash$t}\texttt{[Res]}\end{tabular} \\ \hline
\textbf{2} &
  \begin{tabular}[c]{@{}c@{}}Which categories do you think sentence: \textcolor[rgb]{1,0,0}{$\backslash$n} \texttt{[SENT]} \textcolor[rgb]{1,0,0}{$\backslash$n} belongs\\ to, out of \texttt{[CATE]}? \textcolor[rgb]{1,0,0}{$\backslash$n} \texttt{[Res]}\end{tabular} \\ \hline
\textbf{3}   & ......                      \\ \hline
\end{tabular}}
\end{table}

\begin{figure}[h]
\centering
\includegraphics[width=\linewidth]{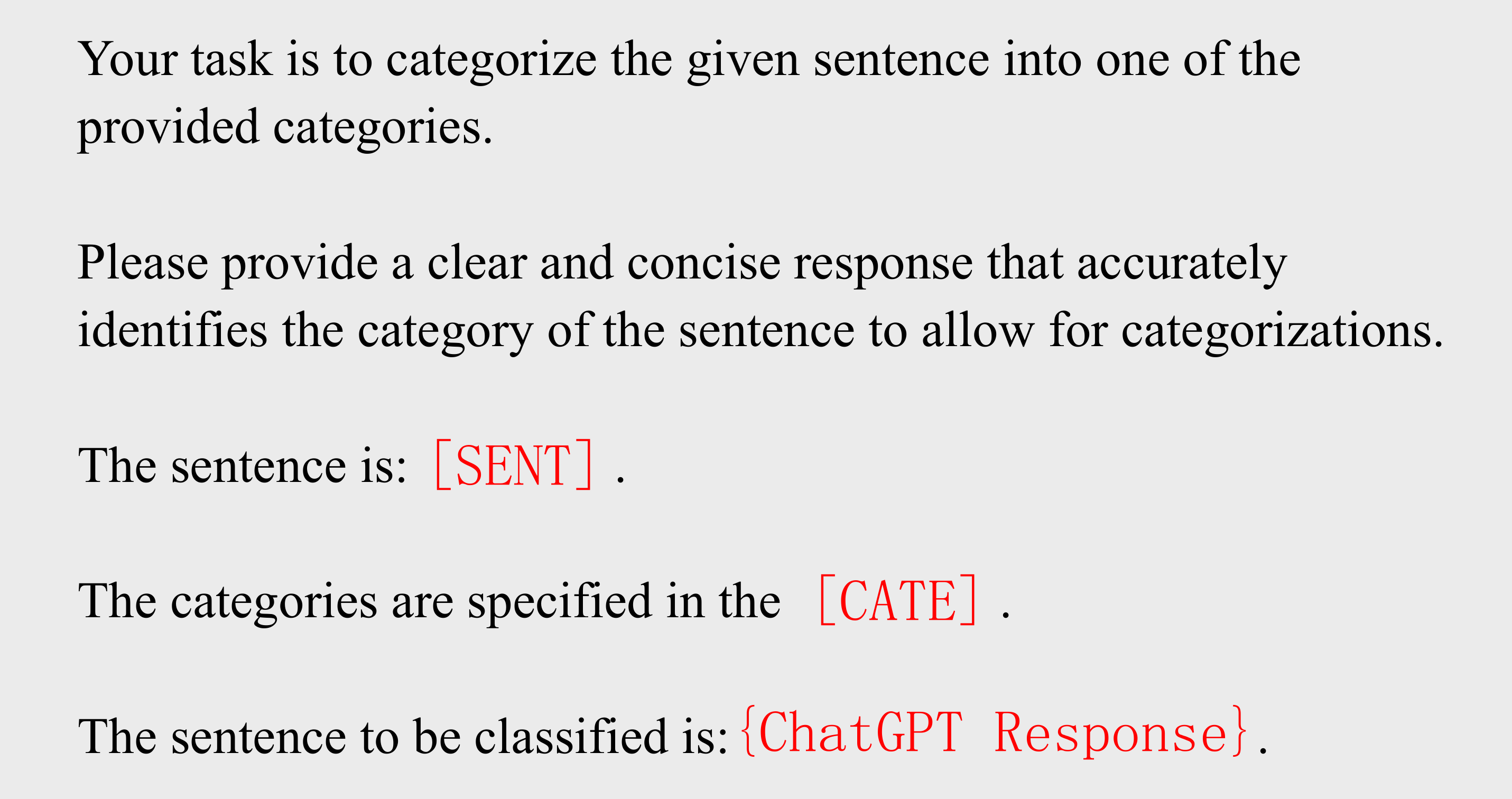}
\caption{The adopted prompt which is selected through the subset evaluation.}
\label{fig_manuallyprompt}
\end{figure}

To conduct the successive comparison experiments, we evaluate the specific effect of each candidate prompt to select the best candidate prompt. Formally, we employ a data sampling-based evaluation approach among these candidate prompts \cite{ChatGPT_code}. Concretely, we randomly selected a fixed number of samples (set as 100 during experiments by default) from the Twitter Natural Hazards dataset, then we further test the performance for each prompts on this subset by accuracy. After overall comparisons, the prompt which is shown in Fig.~\ref{fig_manuallyprompt} is selected as the most suitable manually defined prompt for subsequent experiments.  

Moreover, note that we add an extra command ``\textit{Please only answer the category.}'' into prompts to ask ChatGPT not to generate redundant explanation around the ChatGPT reply, which might be a disrupting factors for subsequent text label decisions. The factor has also been taken into consideration for the subsequent prompting methods.

\subsubsection{ChatGPT Triggered Prompts}
Drawing inspiration from the relevant literature \cite{trans_chatgpt, comparabert_ChatGPT}, we posit that inquiring about ChatGPT itself could potentially yield valuable insights into the generation of high-quality templates. Thus, we seek inspiration from ChatGPT by asking ChatGPT with the recommendations for templates generation. 
Note that a similar preliminary study of Zhong \textit{et al.} \cite{comparabert_ChatGPT} suggests that the task-specific prompts can be triggered by using the following human inquiries:

\begin{tcolorbox}[notitle,boxrule=0pt,colback=gray!20,colframe=gray!20]
\texttt{> Provide five concise prompts or templates that can make you deal with the \textbf{[x]} task.}
\end{tcolorbox}

where the slot \texttt{[x]} means the specific task types. Experiments prove that this strategies performs well in most scenarios.  

\begin{figure}[h]
\centering
\includegraphics[width=\linewidth]{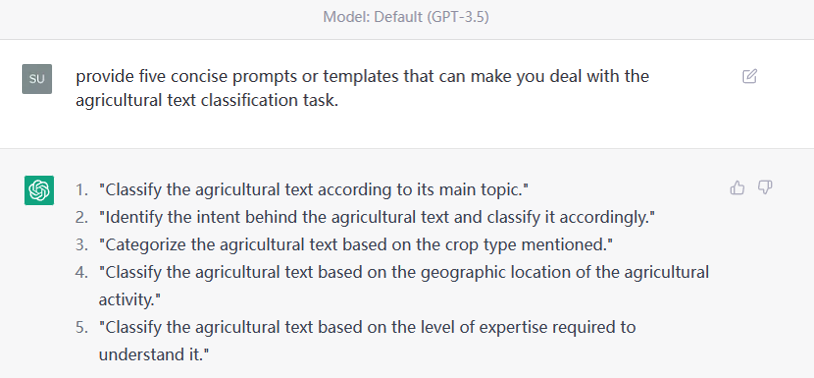}
\caption{Candidate prompt templates triggered by requests to ChatGPT (Model: GPT-3.5, Query Date: 2023.4.02).}
\label{fig_promptsofChatGPT}
\end{figure}

Correspondingly, as shown in Fig.~\ref{fig_promptsofChatGPT}, our request is intuitively constructed as follows:

\begin{tcolorbox}[notitle,boxrule=0pt,colback=gray!20,colframe=gray!20]
\texttt{> Provide five concise prompts or templates that can make you deal with the \textbf{agricultural text classification} task.}
\end{tcolorbox}

Afterwards, ChatGPT naturally answers us with several candidate responses, which is depicted in Fig.~\ref{fig_promptsofChatGPT}. 
The prompts that have been generated appear to be sensible and consistent in terms of their semantic content, while also exhibiting some noticeable distinctions in terms of their individual formats.

To this end, following the above described sampling-based evaluation method, we select the best-performed prompt to represent the ChatGPT triggered prompts for successive comparison experiments, which is shown as follows: 

\begin{tcolorbox}[notitle,boxrule=0pt,colback=gray!20,colframe=gray!20]
``\texttt{> Classify the agricultural text: \textbf{[SENT]} according to its main topic \textbf{[CATE]}.}''
\end{tcolorbox}

\subsubsection{Zero-Shot Similarity Prompts}
Motivated by previous few/zero-shot learning works that utilizes meta-learning paradigm \cite{fewshotsurvey,IE_ChatGPT}, we devised a novel prompting strategies upon it, called zero-shot similarity-based prompting. 

Typically, few-shot object classification is performed by leveraging sample and classifiers from similar classes by some distance measure and similarity functions, such as cosine similarity and squared $\ell_{2}$ distance \cite{fewshotsurvey}. To give an example, let's consider the few-shot learning-based images classification task. Firstly, given an image to be classified, one extra representative image for each category was choosed. Then, they were embedded into the same low-dimensional space using an embedding network, such as siamese network, prototypical network, and matching network. Finally, the similarity threshold between the image to be classified and images from all-kind of categories is then used for label classification.

Back to agricultural text classification, the adopted ChatGPT interface can be regarded as a special distance similarity measurement for evaluating the inter-relationship between two different sentences. All these procedures were conducted by performing one turn or multi turns QA. Specifically, we have designed two QA modes: end-to-end direct QA-based similarity evaluation and progressive comparison QAs-based similarity evaluation. 

\begin{itemize}

\item \textbf{End-to-end direct QA-based:} Concretely, the most straightforward and simplest way is to directly ask ChatGPT that which sentence is most similar to the pre-classified sentence.
Furthermore, we adopt the following prompt during experiments.

\begin{tcolorbox}[notitle,boxrule=0pt,colback=gray!20,colframe=gray!20]
\texttt{> Given sentence S:\textbf{[SENT1]}, which sentence of A:\textbf{[SENT2]}, B: \textbf{[SENT3]}, ... do you think is most similar to sentence S? A, B, ..., or C?}
\end{tcolorbox}

In this manner, the text category can be finally determined. As see in Fig.~\ref{fig_ChatGPTend}, the target sentence can be classified to the category of sentence \textbf{C} based on only one-turn QA.

\begin{figure}[h]
\centering
\includegraphics[width=\linewidth]{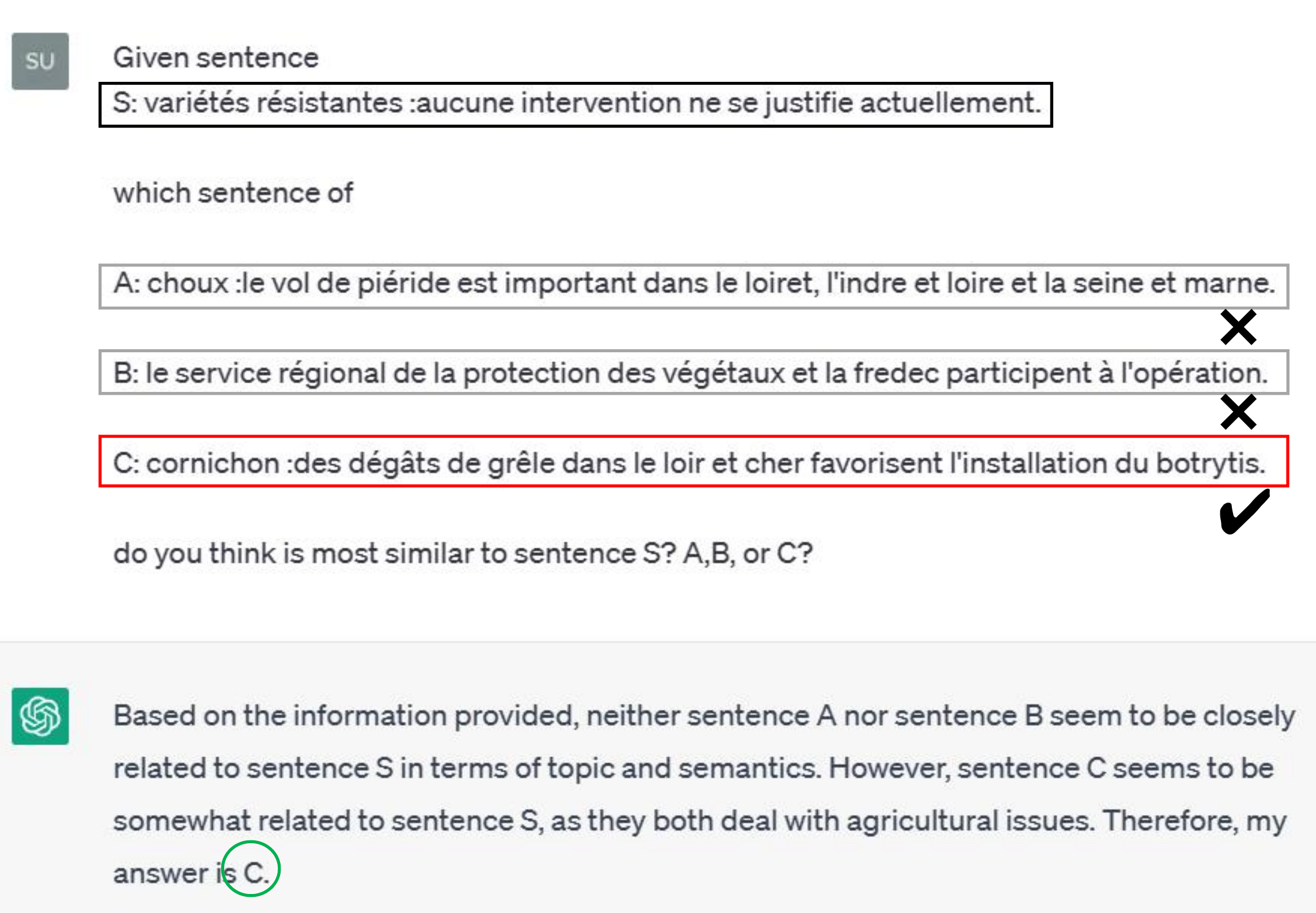}
\caption{The end-to-end direct similarity measurement QA-based prompting method for text classification.}
\label{fig_ChatGPTend}
\end{figure}

\item \textbf{Progressive comparison QAs-based:} Similar to bubble sorting algorithm that compares pairs of elements at a time and subsequently applying the comparison to successive elements. Encouraged by the sorting algorithm, we incorporate its use in determining text similarity. Intuitively, we use the QA prompt:

\begin{tcolorbox}[notitle,boxrule=0pt,colback=gray!20,colframe=gray!20]
``\texttt{> Given sentence S: \textbf{[SENT0]}, which sentence A: \textbf{[SENT1]} and B: \textbf{[SENT2]} do you think is more similar to sentence S?} \\ \texttt{\underline{Please answer using only A and B.}}''.
\end{tcolorbox}

\begin{figure}[h]
\centering
\includegraphics[width=\linewidth]{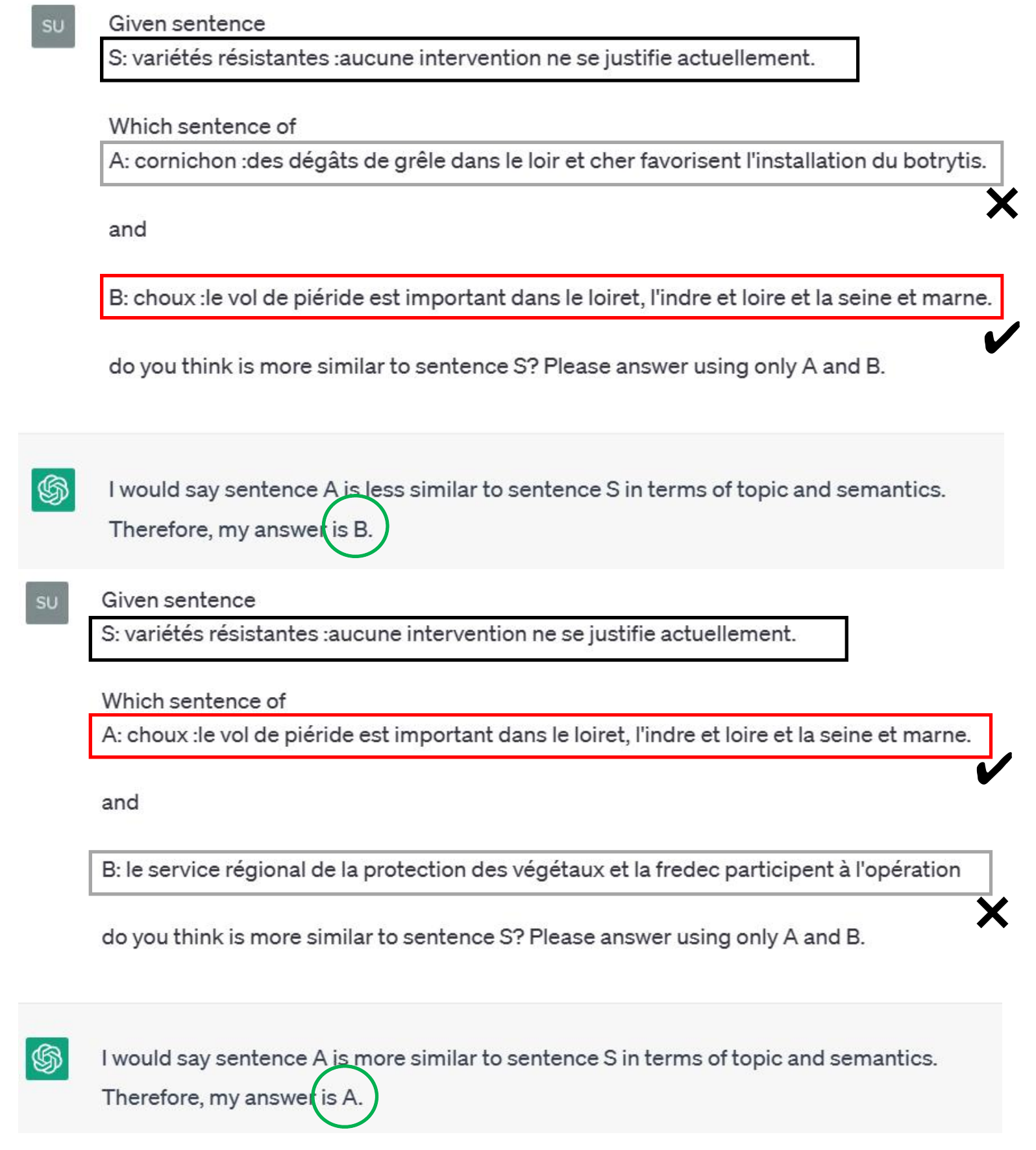}
\caption{The progressive similarity measurement QAs-based prompting method for text classification.}
\label{fig_ChatGPTbubble}
\end{figure}

A typical example related to the three-classification problem was given in Fig.~\ref{fig_ChatGPTbubble}. Based on two-turn QAs, the target sentence can be classified to the category of sentence \textbf{A} based on the topic similarity comparison in the second QA stage. To our knowledge, we are the first to utilize the multi-stage similarity comparison approach to conduct the text classification task.
\end{itemize}

\subsubsection{Chain-of-Thought Triggered Prompts}
In Jiao \textit{et al.}s' \cite{trans_chatgpt} preliminary research of ChatGPT evaluation, they devised a \textit{Pivot Prompting} translation strategy for ChatGPT-based multi-linguistic translator, which significantly improves the translation performance. \textit{Pivot Prompting} translates source language to target language by using a high-resource pivot language (i.e. English by default) as a transition when two distant language is scarce. The above research reflected that this intermediate transitional strategy is particularly effective in some special application scenarios. Jin \textit{et al.}'s knowledge graph-based QA research \cite{jwqrcekbqa} provides further evidence that these chains of reasoning are a critical factor that impacts the accuracy of the model. 

\begin{figure}[h]
\centering
\includegraphics[width=\linewidth]{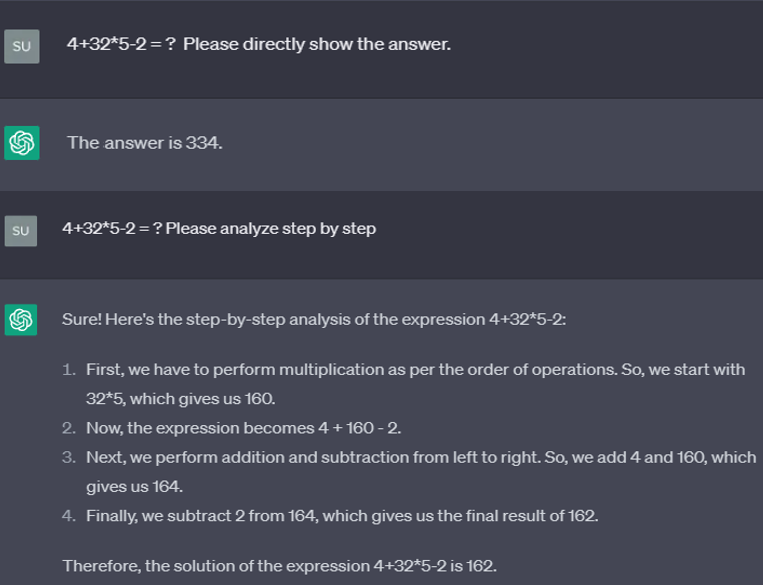}
\caption{The ChatGPT performance comparison between providing the answer directly and presenting a step-by-step calculation process in solving arithmetic problems. (Model: GPT-3.5, Query Date: 2023.3.15)}
\label{fig_ChatGPTcalcul}
\end{figure}

Moreover, our inspection of ChatGPT's computational ability reveals that while ChatGPT tends to fall behind in its ability to reason and provide correct answers, it performs competitively when a step-by-step calculation process is used. Fig.~\ref{fig_ChatGPTcalcul} gives a typical example. To be more specific, while ChatGPT incorrectly provides the answer of 334 for the arithmetic problem 4+32 $\cdot$ 5-2, it is capable of correctly reasoning and arriving at the right answer for the same problem based on a step-by-step calculation process.

\begin{figure}[h]
\centering
\includegraphics[width=\linewidth]{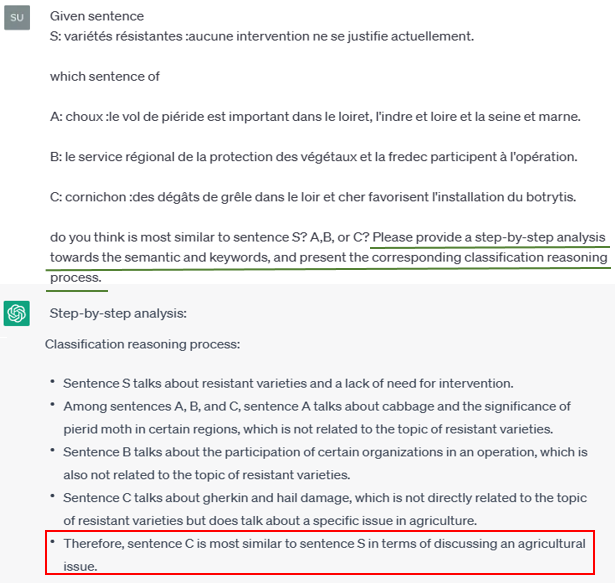}
\caption{The \textit{Chain-of-Thought}-based prompting strategy which is built upon a simple and direct QA prompt. (Model: GPT-3.5, Query Date: 2023.3.15)}
\label{fig_ChatGPTstepbystep}
\end{figure}

Building upon the experimental findings that support the effectiveness of step-by-step incremental reasoning, we explore the utility and viability of utilizing this technique for agricultural text classification. Concretely, we choose the \textit{manually defined prompts} and \textit{ChatGPT triggered prompts} as baselines. Also, drawing on these initial prompts, we require ChatGPT not only with delivering the final classification category but also with producing a corresponding comprehensive Chain-of-Thought reasoning analysis. For ease of illustration, as shown in Fig.~\ref{fig_ChatGPTstepbystep}, we further add the following expression based on the original QA prompt.

\begin{tcolorbox}[notitle,boxrule=0pt,colback=gray!20,colframe=gray!20]
 ``\texttt{> Please provide a step-by-step analysis towards the semantic and keywords, and present the corresponding classification reasoning process.}''
 \end{tcolorbox}
 
This section presents several feasible strategies which shares a diverse range of distinctive features that set them apart from one another. But the prompting engineering is more complex and nuanced than what we can observe at a superficial level, as it is influenced by multiple factors, with dataset features playing a particularly significant role. For example, experimental results indicated that the Chain-of-Though triggered prompts performs particularly well on datasets with a high number of classification categories, but its effectiveness is not satisfactory when dealing with datasets with relatively simple classification (few categories), such as only two to three categories. 

The upcoming experiments will systematically compare multiple prompting strategies proposed above to enable a comprehensive evaluation and research.


\subsection{ChatGPT Q\&A Inference}
ChatGPT is a state-of-the-art conversation robot which are based on the generative language model, Generative Pre-trained Transformer (GPT). The ChatGPT model's conversational capability stems from its ability to generate coherent text using sequence-to-sequence learning and the transformer architecture, where it conditions on a given prompt and samples from a probability distribution of words. The prominent intelligent thinking of ChatGPT is derived from its training on extensive amounts of text data to acquire a statistical understanding of the patterns that exist in natural language.

The GPTs family uses the transformer architecture, which is a deep neural network that processes input data in parallel using multi-headed attention mechanisms. During the inference stage, the GPT model generates text by conditioning on a given prompt and sampling from a probability distribution of words that follow. The probability distribution is computed by applying the softmax function over the output of the model. The output of the model at each time step depends on the previous tokens generated, creating a generative process that allows the model to generate coherent text.

Mathematically, the token generative procedure of ChatGPT can be represented as:

\begin{equation}
\label{gptreason}
p(y|x) = \prod_{t=1}^{T} p(y_{t}|y_{1},...,y_{t-1}, x) 
\end{equation}

where the $\prod$ means the probability multiplication operator. Given the previous tokens $y_{1},...,y_{t-1}$ and the input prompt $x$, $p(y_{t}|y_{1},...,y_{t-1}, x)$ is the probability distribution over the token $y_{t}$ in t-th time step and $T$ is the length of the generated sequence.

At this stage, we direct our focus towards ChatGPT and hypothesize that ChatGPT possesses inherent capabilities that enable it to act as an integrated zero-shot text classification interface through an interactive mode.

During the ChatGPT interaction process, we created a fresh conversation thread for each prompt to ensure that the previous conversation history would not impact ChatGPT's responses. By adopting this methodology, ChatGPT is able to consistently exercise independent thinking and deliver optimal responses by leveraging the information provided by the user.

Besides applying the vanilla ChatGPT (GPT-3.5), our experiments also evaluated the capabilities of GPT-4 \cite{GPT4}. GPT-4 represents a new breakthrough in OpenAI's ongoing efforts to advance the field of deep learning. The results showed that GPT-4 performed better than ChatGPT, even in some complex semantic text classification scenarios, as seen in the following section of related evaluations.

\subsection{Answer Alignment}
After the above steps, using an appropriate prompt and ChatGPT for question-answering, ChatGPT provided feedback on the classification results for the corresponding text. Nevertheless, its unique characteristic of generating responses in a conversational way presents challenges for the subsequent analysis and evaluation of its outputs. Unlike traditional PLM-based text classification models, ChatGPT's responses do not directly correspond to predefined labels, which means that an additional alignment strategy is required to convert these intermediate answers into the final labels that can be used to calculate various performance metrics (e.g. accuracy and F1-score). We refer to this additional mapping strategy as the ``answer alignment engineering''.

In our experiments, we investigated the impact of answer alignment engineering on the ChatGPT-based text classification's performance. Specifically, we designed and implemented two different alignment strategies: rule-based matching strategy and similarity-based matching strategy. Both approaches involve a mapping process that maps the intermediate responses to the corresponding labels. The rule-based matching approach uses predefined rules to match the responses to the labels, while the string matching approach computes the similarity between the response and each label and selects the label with the highest similarity score.

\begin{figure}[h]
\centering
\includegraphics[width=\linewidth]{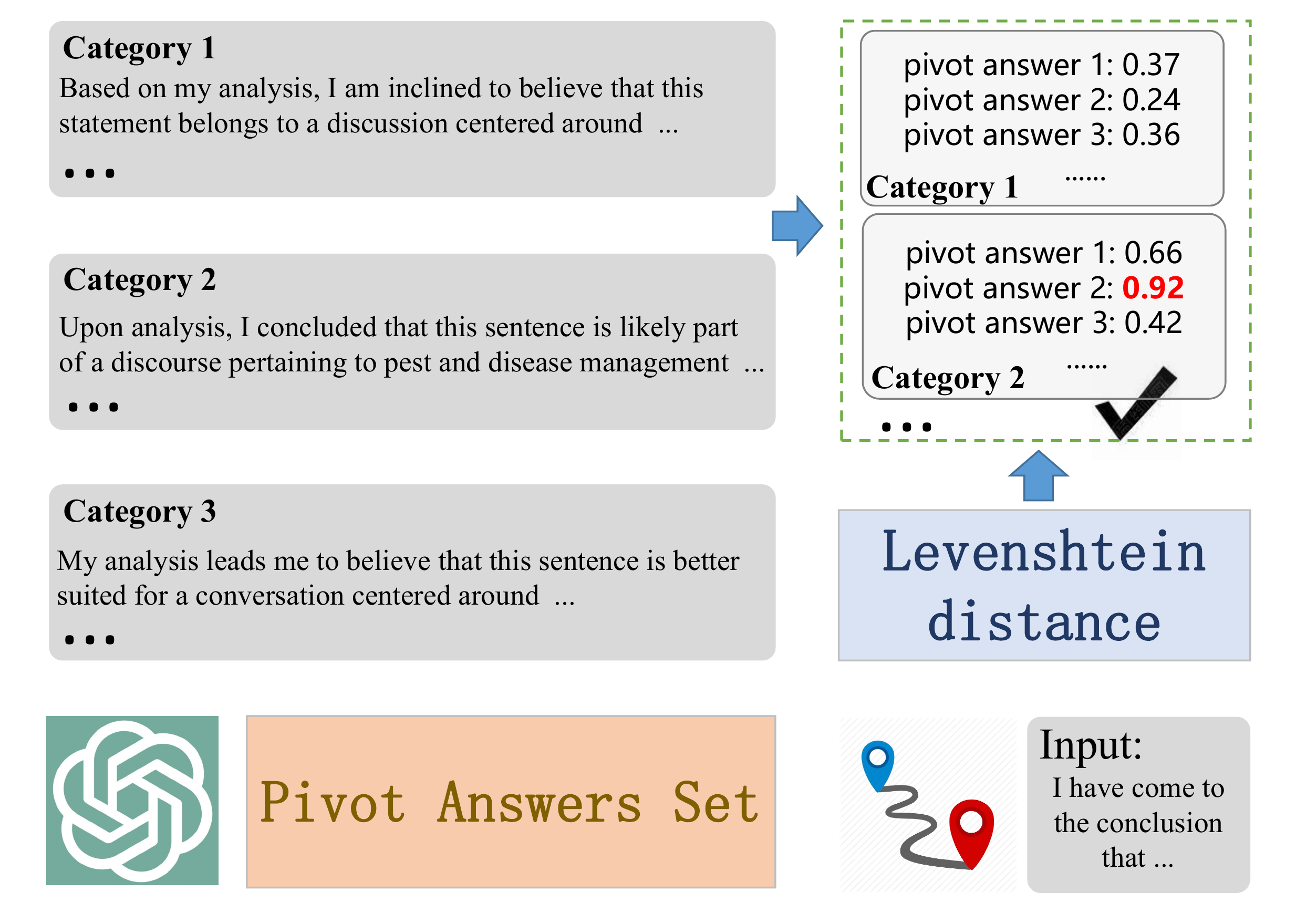}
\caption{The illustrating diagram of the similarity-based matching strategy.}
\label{fig_similar_strat}
\end{figure}

\begin{itemize}
\item \textbf{Rule-based matching strategy:} 
Essentially, the rule-based matching strategy is a text matching method that involves using patterns or rules based on token attributes, such as part-of-speech tags, to match sequences of tokens in unstructured text data. During our experiments, we use the Matcher\footnote{\url{The Matcher tool is in https://spacy.io/api/matcher} [Accessed on 2023.03] } object in spaCy v3 to find the matched tokens in context to classify the sentence returned by ChatGPT. spaCy v3 is a leading industrial-strength natural language processing and analysis tool\footnote{Spacy can be accessed on \url{https://spacy.io} [Accessed on 2023.03].} using Python. 

Specifically, we firstly analyze the text extraction patterns based on expert experience and ChatGPT's historical output habits, and design and define a set of rules. Then, the rules are applied to the text data and the extracted information is verified and validated. Finally, after adjustment and optimization, a comprehensive set of matching rules is summarized;

\item \textbf{Similarity-based matching strategy:} 
Although the former approach utilizes rigid matching with high accuracy, it is difficult to handle semantically ambiguous situations. To address this issue, we adopt the second strategy, which is the similarity-based matching strategy.
Firstly, we aggregate and synthesize ChatGPT's commonly expressed utterances under each category to establish a repository of pivot answers for each category. Subsequently, we apply the Levenshtein distance algorithm to compute the minimum edit distance between each pivot answer and the input answer being classified. The pivot answer with the smallest edit distance is regarded as the definitive category label. This approach offers comprehensive coverage and effectively mitigates the shortcomings of rule-based matching in accommodating ambiguous and nuanced language use.

The string similarity-based matching strategy is depicted in Fig.~\ref{fig_similar_strat}.
\end{itemize}

In theory, neither of these two strategies can perfectly solve the problem of answer mapping. To overcome the challenge of answer mapping, we combined rule-based and similarity-based matching strategies in a pipeline approach. Specifically, we found that ChatGPT typically provides explicit category labels in natural language form. Therefore, in the first step, we tend to use the rule-based strategy to parse the intermediate answers. If the category is still uncertain, we then use the string similarity-based strategy to compute the similarity between the intermediate answer and each category's answer examples, selecting the category with the highest similarity as the final classification. In our experiments, this approach can simultaneously improve the accuracy and recall rate of the answer mapping process effectively.

Nevertheless, this work mainly explored a character-based literal matching method that lacks semantic understanding. The method has certain limitations, whereas the deep neural network-based methods using PLMs are more adept at such scenarios. In our future work, we will attempt to use a PLMs-based semantic understanding model for this step, which theoretically can bring about better performance.

\section{Experimental Setup}
\label{experimentssetup}
We perform a series of experiments in order to figure out exactly what kinds of factors of these devised strategies that indeed influence the final agricultural text classification performance of the ChatAgri in Section~\ref{experiments}. Correspondingly, acting as a preliminary, this section mainly introduces the details of the experimental setups, including the used multi-linguistic datasets, the employed text classification baselines for model comparisons, the adopted evaluation metrics, and the adopted hyperparameters of our ChatAgri.

\subsection{Datasets}
To demonstrate the actual potentials of ChatAgri for classifying agricultural text, we carefully collect several suitable datasets for evaluation and validation, ranging from different types of categories (e.g. plant diseases, insect pests, and twitter natural hazards) and numbers of categories to different languages, including French, English, and Chinese. These datasets are respectively called Amazon-Food-Comments, PestObserver-France, Natural-Hazards-Twitter, and Agri-News-Chinese in our experiments, whose details are illustrated as follows.

\begin{table*}[h]
\renewcommand{\arraystretch}{1.3}
\caption{The statistical meta information of the adopted agricultural text classification datasets. \label{table_datesets}}
\centering
\scalebox{0.9}{
\begin{tabular}{cccccc}
\hline
\textbf{Dataset}     & train samples & test samples & language & categories                         & label count \\ \hline
Amazon-Food-Comments & 165863        & 16175        & English  & 'negative', 'positive', 'neutral'  & 3           \\
PestObserver-France  & 322           & 80           & French   & 'Bioagressor', 'Disease', 'Others' & 3           \\
Natural-Hazards-Twitter & 45669         & 5074         & English  & 'negative', 'positive'             & 2           \\
Natural-Hazards-Type & 5000  & 1000 & English & \begin{tabular}[c]{@{}c@{}}'Hurricane', 'Wildfires', 'Blizzard',\\  'Floods', 'Tornado'\end{tabular} & 5 \\
Agri-News-Chinese       & 52000 & 6500 & Chinese & 
\begin{tabular}[c]{@{}c@{}}'Agricultural economy', 'Horticulture',\\  'Agricultural engineering', Farming', \\ 'Fisheries','Forestry','Crops'\end{tabular}
& 7 \\ \hline
\end{tabular}}
\end{table*}


\begin{itemize}
\item \textbf{Amazon-Food-Comments:} An amazon food comment dataset that contains nearly 200,000 positive samples, neutral samples, and negative samples, which can be used to perform text classification tasks for both positive, neutral, and negative reviews\footnote{Access to \url{https://nijianmo.github.io/amazon/index.html} for more details of Amazon-Food-Comments 
[Accessed on 2023.02].};
\item \textbf{PestObserver-France: \cite{planthealthbulletin}} A plant health bulletin classification dataset in French to estimate a agricultural prediction model that how well can it deal with  heterogeneous documents and predict for natural hazards\footnote{PestObserver-France can be downloaded from \url{https://github.com/sufianj/fast-camembert} [Accessed on 2023.02].};
\item \textbf{Natural-Hazards-Twitter: \cite{nathaztwitter}} A natural disaster dataset with sentiment labels of United States which is proposed to identify attitudes towards disaster response. It contains different natural disaster types and nearly 5,000 Twitter sentences\footnote{Natural-Hazards-Twitter can be downloaded from https://git- \url{hub.com/Dong-UTIL/Natural-Hazards-Twitter-Dataset} [Accessed on 2023.02].};
\item \textbf{Natural-Hazards-Type:} In addition to recognize the sentiment polarities of Natural-Hazards-Twitter, we also re-organize it into a new disaster type classification dataset, denoted as Natural-Hazards-Type, to identify the natural disaster categories of text. Due to the large volume of the original Natural-Hazards-Twitter dataset, the new Natural-Hazards-Type dataset has taken a small subset of it, containing thousands of samples; 
\item \textbf{Agri-News-Chinese:} Besides the above existing datasets, we proposed a Chinese Agricultural short text classification dataset, namely Agri-News-Chinese, containing seven categories, such as agricultural economy and aquatic fishery. Its data source was collected and cleaned from the agricultural technology expert online system (ATE expert online system) \footnote{More details about ATE expert online system can be accessed to \url{http://zjzx. cnki.net/} [Accessed on 2023.02].}, with a total volume of approximately 60000 pieces of data, divided into the train and test sets by 9:1.
\end{itemize}

Table~\ref{table_datesets} gives a meta statistic for the five datasets, including the split distribution of train/test samples, the language scope, and the categories of textual topics.

\subsection{Baselines} 
Existing extensive models for text classification can be divided into five major training paradigms: 1) traditional feature engineering-based machine learning (e.g. SVM, Decision Tree, and Random Forest) \cite{agriieee,foodsecurity,ieeeaccdff}; 2) word embedding-based deep learning (e.g. TextCNN, and TextRNN); 3) PLM-based fine-tuning, in which the PLMs include BERT \cite{BERT}, BART \cite{BART}, T5 \cite{T5} and so on; 4) PLM-based prompt learning; and 5) the newest ChatGPT QA-based zero-shot learning paradigm that brought by ChatGPT recently (e.g. ChatIE \cite{IE_ChatGPT}, ChatEventExtract \cite{eventextr_ChatGPT}, and our ChatAgri).

To ensure the research comprehensiveness, the above introduced mainstream natural language understanding (NLU) paradigms were considered to be estimated and reported as the comparison baselines in our experiments. Specifically, besides the herein proposed ChatAgri, we adopted the following methods listed below for each mentioned learning paradigm.

\begin{itemize}
\item \textbf{SVM: \cite{agriieee}} Support Vector Machine (SVM) is a classic classification method pursuing maximization of support vector distance between multiple class hyper-planes for classification, typically in the text category classification task. SVM mainly classifies the text by calculating the unstructured discrete textual features, optimizing them into high-dimensional spatialized vector representations;
\item \textbf{Random Forest: \cite{agriieee}} Random Forest (RF) is also a well-known  classification algorithm, belonging to the ensemble methods family, combines multiple weaker classifier to create a stronger classifier for categorical data;
\item \textbf{TextCNN: \cite{kimtextcnn}} Built on the top of pre-trained word vectors, TextCNN uses convolutional neural networks (CNN) as feature detector and utilizes kernels of different sizes to extract the valuable semantic feature for sentence classification. Lastly, the external softmax layer performs multi-classification on the convolutional logical values;
\item \textbf{TextRNN: \cite{lpftextrnn}} Based on pre-trained word embeddings, TextRNN integrates recurrent neural network (RNN) into the multi-learning framework. Specifically, TextRNN utilizes long short-term memory (LSTM) to address the issues of gradient vanishing and exploding, thereby resolving the challenge of capturing long-range dependencies within sequences;
\item \textbf{BERT-based fine-tuning: \cite{BERT,BERT_agri_sentiana,survey_ChatGPT}} Fine-tuning BERT has emerged as a widely employed methodology across diverse text processing tasks, including text classification. By generating contextualized word embeddings, BERT effectively captures both semantic and syntactic information associated with individual words. Leveraging its inherent strengths, BERT can be fine-tuned on specific tasks utilizing limited labeled datasets, rendering it a flexible and formidable solution for addressing an array of text processing objectives;
\item \textbf{T5-based prompt-tuning: \cite{promptsurvey,openprompt,T5}} Different from the ``pre-train then fine-tune'' procedure of fine-tuning methods, the prompt-tuning paradigm induces those PLM to generate suitable target responses with the help of additional triggered sentences, which are called ``prompts''. In prompt-tuning, the major research attention has been transferred on how to provide better prompts to activate the PLM's rich internal prior knowledge. We use the PLM, Transfer Text-to-Text Transforme (T5) to be the backbone. T5 is a unified very large PLM based on Transformer architecture, which converts all text processing tasks into Text-to-Text tasks; 
\item \textbf{BART-based prompt-tuning: \cite{openprompt,BART}} We also investigate the usage of Bidirectional and Auto-Regressive Transformers (BART), being acted as the backbone for prompt learning. BART simultaneously incorporates the advantages of BERT and GPT (i.e. the characteristics of the context bidirectional modelling and the sequence joint probability hypothesis);
\end{itemize}

\subsection{Evaluation Metrics}
In such agricultural text classification task that involves multiple label classification, accuracy and F1-score are two commonly used metrics.

Correspondingly, accuracy measures the proportion of correctly predicted samples among all predicted samples, is a simple and coarse-grained evaluation metric which only accumulates all the correct instances. And accuracy is calculated as follows:
\begin{equation}
\label{eq_acc}
Accuracy = Count_{T}/Count_{N}
\end{equation}
where $Count_{T}$ represents the correctly predicted samples and $Count_{N}$ represents the total number of samples evaluated.

Comparatively, F1-score is considered to be a relatively fine-grained evaluation indicator than accuracy. In comparison to accuracy, F1-score is considered to be a higher confidence indicators which simultaneously considers the precision and recall. And F1-score is calculated as follows:

\begin{equation}
\label{eq_f1_score}
\begin{array}{l}
F 1=\displaystyle \frac{2 * \text{Precision} * \text{Recall}}{\text{Precision}+ \text{Recall}} \quad \text { where } \quad \\
\\
\text{Precision}=\displaystyle \frac{T P}{T P+F P} \quad \& \quad \text{Recall}=\displaystyle \frac{T P}{T P+F N}
\end{array}
\end{equation}

In the equation presented, $Precision$ and $Recall$ refer to the precision and recall rate of the classification results, respectively. $TP$ (true positives) represents the number of samples whose actual and predicted class are both positive; $FP$ (false positives) represents the number of samples whose actual class is negative but are predicted as positive; and $FN$ (false negatives) represents the number of samples whose actual class is positive but are predicted as negative.

Specifically, F1-score includes several calculating strategies: micro-F1, macro-F1, and weighted-F1. Without considering micro-F1 and macro-F1, we utilize the weighted-F1 as it accounts for the classification performance of categories under varying weights, thereby providing greater reference value.

\subsection{Hyperparameter Settings}
During our experimental procedure, there are various meta settings for all kinds of hyperparameters. The optimal hyperparameters, determined by their superior performance on the development set, will be selected for the final evaluation. The meta settings are summarized as follows.

We adopted the pretrained word vectors, GloVe \cite{GloVe}, as the embeddings of the baselines of TextCNN and TextRNN. GloVe leverages the word co-occurrence statistics that can capture both syntactic and semantic relationships between words$\footnote{GloVe embedding can be downloaded from: \url{https://nlp.stanford.edu/projects/glove/} [Accessed on 2023.03]. }$. 
Considering the trade-offs between computational limitations and performances and to ensure experimental competitiveness and stability, we adopted the version ``bert-base-uncased''$\footnote{BERT can be obtained from: \url{https://huggingface.co/docs/transformers/model_doc/bert} [Accessed on 2023.03]. }$ for the PLM BERT, the version ``t5-base''$\footnote{T5-base can be obtained from: \url{https://huggingface.co/t5-base} [Accessed on 2023.03]. }$ for the PLM T5, and the version ``facebook/bart-base''$\footnote{BART can be obtained from: \url{https://huggingface.co/docs/transformers/model_doc/bart} [Accessed on 2023.03]. }$ for the PLM BART respectively.
The code implementation is developed using Python 3.7 $\footnote{Python can be downloaded from: \url{https://www.python.org/downloads/release/python-370} [Accessed on 2023.03]. }$ and PyTorch 1.9.0 $\footnote{Pytorch can be downloaded from: \url{https://pytorch.org/blog/pytorch-1.9-released} [Accessed on 2023.03]. }$ frameworks. 
For experimental simplicity, the prompts of the prompt-tuning baselines are pre-defined as ``\texttt{Given a sentence of [SENT], it is more like to be a topic of \textbf{\{SLOT\}} from [CATE]}'', and the probability scores of the estimated words in the position of \texttt{\textbf{\{SLOT\}}} are then regarded as the intermediate answers for the final classification.
Furthermore, the experimental hardware environment comprises a CPU Intel Core i9-9900k, and a single Nvidia GPU of \textit{GTX 1080Ti}.

\section{Experimental Results and Analyses}
\label{experiments}
Next, we conducted a series of baseline comparison experiments and ablation experiments to analyze and explore the specific connections between various key factors that affect the performance of ChatAgri on agricultural text classification tasks. We first verified the competitiveness and superiority of ChatAgri relative to known state-of-the-art (SOTA) models. Then, we systematically investigated the impact of different prompting strategies on the classification accuracy for text classification. Moreover, we also attempted to apply GPT-4 and investigated the superiority of GPT-4 compared to the basic version of ChatGPT, GPT3.5. The systematic analysis toward extensive empirical results firmly demonstrate the enormous potentials, feasibility, and broad application prospects of ChatGPT in agricultural text classification tasks. 

\begin{table*}[h]
\renewcommand{\arraystretch}{1.3}
\caption{Performance Statistics of all baselines and ChatAgri on all adopted datasets. We respectively boldface and underline the score with the best performance and the second-best performance across all models \textbf{(Query Date: 2023.3.16)}. \label{table_comparisons}}
\centering
\scalebox{0.8}{
\begin{tabular}{cccccccccccc}
\hline
 &
   &
  \multicolumn{2}{c}{\begin{tabular}[c]{@{}c@{}}Amazon-Food\\ -Comments\end{tabular}} &
  \multicolumn{2}{c}{\begin{tabular}[c]{@{}c@{}}PestObserver\\ -France\end{tabular}} &
  \multicolumn{2}{c}{\begin{tabular}[c]{@{}c@{}}Natural-Hazards\\ -Twitter\end{tabular}} &
  \multicolumn{2}{c}{\begin{tabular}[c]{@{}c@{}}Natural-Hazards\\ -Type\end{tabular}} &
  \multicolumn{2}{c}{\begin{tabular}[c]{@{}c@{}}Agri-News\\ -Chinese\end{tabular}} \\ \cline{3-12} 
\multirow{-2}{*}{\textbf{\begin{tabular}[c]{@{}c@{}}Learning\\ Paradigms\end{tabular}}} &
  \multirow{-2}{*}{\textbf{\begin{tabular}[c]{@{}c@{}}Baseline\\ Methods\end{tabular}}} &
  acc &
  \begin{tabular}[c]{@{}c@{}}weighted\\ -F1\end{tabular} &
  acc &
  \begin{tabular}[c]{@{}c@{}}weighted\\ -F1\end{tabular} &
  acc &
  \begin{tabular}[c]{@{}c@{}}weighted\\ -F1\end{tabular} &
  acc &
  \begin{tabular}[c]{@{}c@{}}weighted\\ -F1\end{tabular} &
  acc &
  \begin{tabular}[c]{@{}c@{}}weighted\\ -F1\end{tabular} \\ \hline
 &
  \textbf{SVM} &
  0.627 &
  0.624 &
  0.672 &
  0.655 &
  0.763 &
  0.742 &
  0.811 &
  0.811 &
  0.523 &
  0.522 \\
\multirow{-2}{*}{\begin{tabular}[c]{@{}c@{}}Traditional\\ Machine Learning\end{tabular}} &
  \textbf{Random Forest} &
  0.647 &
  0.643 &
  0.664 &
  0.652 &
  0.787 &
  0.755 &
  0.863 &
  0.863 &
  0.553 &
  0.534 \\ \hline
 &
  \textbf{TextCNN} &
  0.748 &
  0.742 &
  0.715 &
  0.704 &
  0.834 &
  0.816 &
  0.914 &
  0.914 &
  0.792 &
  0.785 \\
\multirow{-2}{*}{\begin{tabular}[c]{@{}c@{}}Word Embedding\\ -based learning\end{tabular}} &
  \textbf{TextRNN} &
  0.727 &
  0.725 &
  0.707 &
  0.697 &
  0.845 &
  0.827 &
  0.931 &
  0.931 &
  0.812 &
  0.801 \\ \hline
\begin{tabular}[c]{@{}c@{}}PLM-based\\  fine-tuning\end{tabular} &
  \textbf{\begin{tabular}[c]{@{}c@{}}BERT-based \\ fine-tuning\end{tabular}} &
  0.767 &
  0.764 &
  0.736 &
  0.714 &
  0.869 &
  0.839 &
  0.945 &
  0.945 &
  0.826 &
  0.819 \\ \hline
 &
  \textbf{\begin{tabular}[c]{@{}c@{}}T5-based \\ prompt-tuning\end{tabular}} &
  \textbf{0.805} &
  \textbf{0.798} &
  \underline{0.764} &
  0.753 &
  \underline{0.874} &
  \underline{0.857} &
  0.966 &
  0.966 &
  0.859 &
  0.854 \\ \cline{2-12} 
\multirow{-2}{*}{\begin{tabular}[c]{@{}c@{}}PLM-base \\ prompt-tuning\end{tabular}} &
  \textbf{\begin{tabular}[c]{@{}c@{}}BART-based \\ prompt-tuning\end{tabular}} &
  \underline{0.800} &
  \underline{0.795} &
  0.757 &
  \underline{0.767} &
  \textbf{0.875} &
  \textbf{0.865} &
  \underline{0.971} &
  \underline{0.971} &
  \textbf{0.867} &
  \textbf{0.862} \\ \hline
\rowcolor[HTML]{FFFFFF} 
\begin{tabular}[c]{@{}c@{}}ChatGPT-based\\ Prompt QA\end{tabular} &
  \textbf{\begin{tabular}[c]{@{}c@{}}ChatAgri-base \\ (Ours)\end{tabular}} &
  0.798 &
  0.793 &
  \textbf{0.794} &
  \textbf{0.789} &
  0.866 &
  0.853 &
  \textbf{0.978} &
  \textbf{0.978} &
  \underline{0.863} &
  \underline{0.856} \\ \hline
\end{tabular}}
\end{table*}

\subsection{Methods Comparison}
Table~\ref{table_comparisons} details comprehensive experimental results on the agricultural text classification task for our model ChatAgri and existing state-of-the-art approaches. In this table, as shown by multiple rows before the row data of \textit{ChatGPT-based Prompt QA}, we conducted a systematic evaluation of the classification performance of these baseline models on these five datasets based on the above described hyperparameter settings. The time node of ChatGPT interface calls is March 16, 2023. Subsequent OpenAI official updates may lead to certain performance fluctuations towards the ChatGPT interface. The last row shows the evaluation results of our ChatAgri. For simplicity and clarity, we took the primary designed solution of ChatAgri as the basic model of ChatAgri for comparison. Specifically, we used the manually defined prompts, which is illustrated in Section~\ref{mdp_subsub}, as the prompting template for ChatAgri. And we simultaneously adopted the rule-based and similarity-based text pattern matching strategy for the answer alignment engineering. Correspondingly, we labeled this basic model of ChatAgri as \textbf{ChatAgri-base}.

In Table~\ref{table_comparisons}, we classified all the existing agricultural text classification methods explored in this experiment according to their belonged learning paradigms. Among them, these methods based on fine-tuning PLM and PLM prompt engineering can be seen as the latest optimal benchmark approaches, and are respectively recorded in the last few rows of the table. From the table, it can be clearly observed that our ChatAgri has achieved exciting and competitive performance on some specific datasets, such as PestObserver-France and Natural-Hazards-Type. Not to mention surpassing traditional machine learning methods or word vector-based representation learning methods by an absolute gap of over 10\% to 20\%, which is a noticeable performance margin. Compared with the latest Transformer PLM-based deep learning methods, ChatAgri is also a particularly strong presence, with no loss in accuracy or weighted-f1 compared to these SOTA methods. Specifically, ChatAgri significantly outperformed the PLM-based fine-tuning method represented by fine-tuned BERT by about 3.0\% accuracy on the PestObserver-France dataset, and outperformed the PLM-based prompt-tuning method represented by prompt-tuned BART by approximately 2.2\% weighted-f1 indicator. Similarly, ChatAgri also surpassed the above two state-of-the-art models by 0.6\% accuracy and weighted-f1 indicators on the Natural-Hazards-Type dataset. In addition, the performance of ChatAgri on other datasets is also impressive. For example, it can be seen from the table that the performance of ChatAgri on the Agri-News-Chinese Chinese dataset have significantly surpassed the PLM-based fine-tuning method represented by fine-tuned BERT by about 3.7\% accuracy and 4.7\% weighted-f1 indicator. In addition, ChatAgri's performance is also slightly higher than the PLM-based fine-tuning method represented by prompt-tuned T5 by approximately 0.4\% accuracy and 0.2\% weighted-f1.

In addition, we further explored the reasons why ChatAgri performed more strongly on some datasets but slightly worse than previous SOTA methods on others. By observations from Table~\ref{table_comparisons}, we found that ChatAgri had obvious advantages on two minority language datasets, PestObserver-France and Agri-News-Chinese, but performed poorly on the widely-used English datasets, Amazon-Food-Comment and Natural-Hazards-Twitter. We speculate that this is mainly due to the difference in the scale of large-scale language corpus training for different languages. After comprehensive investigations on latest literature \cite{ChatGPT_code,GPT4,eloundou2023gpts}, we can conclude that ChatGPT excels at handling various cross-linguistic tasks. Unlike previous methods based on traditional PLMs, ChatGPT's learning corpus is totally comprehensive and of high quality, covering the majority of languages spoken in most countries. Moreover, ChatGPT's ultra-large parameter size allows it to memorize and master more linguistic knowledge, not just limited to English. Therefore, in terms of cross-lingual understanding capability, ChatGPT is significantly superior to traditional PLM models (e.g. BERT, RoBERTa, and BART). Correspondingly, traditional PLM models perform poorly on less commonly spoken language datasets, as their learning corpus is far less comprehensive and of lower quality than that of ChatGPT. This probably is the primary factor that allows ChatAgri to perform well on various minority language datasets regardless of these datasets' linguistic characteristics.

\begin{table*}[h]
\renewcommand{\arraystretch}{1.3}
\caption{Comparative experimental results of ChatAgri-base and various model variants of ChatAgri that utilized various advanced prompts, where the ChatAgri-base can be regarded as a basic ChatAgri implementation \textbf{(Query Date: 2023.3.24)}. \label{table_advanceps}}
\centering
\scalebox{0.8}{
\begin{tabular}{ccccccccccc}
\hline
\rowcolor[HTML]{FFFFFF} 
 &
  \multicolumn{2}{c}{\begin{tabular}[c]{@{}c@{}}Amazon-Food\\ -Comments\end{tabular}} &
  \multicolumn{2}{c}{\begin{tabular}[c]{@{}c@{}}PestObserver\\ -France\end{tabular}} &
  \multicolumn{2}{c}{\begin{tabular}[c]{@{}c@{}}Natural-Hazards\\ -Twitter\end{tabular}} &
  \multicolumn{2}{c}{\begin{tabular}[c]{@{}c@{}}Natural-Hazards\\ -Type\end{tabular}} &
  \multicolumn{2}{c}{\begin{tabular}[c]{@{}c@{}}Agri-News\\ -Chinese\end{tabular}} \\ \cline{2-11} 
\rowcolor[HTML]{FFFFFF} 
\multirow{-2}{*}{\textbf{\begin{tabular}[c]{@{}c@{}}Prompting\\ Strategies\end{tabular}}} &
  acc &
  \begin{tabular}[c]{@{}c@{}}weighted\\ -F1\end{tabular} &
  acc &
  \begin{tabular}[c]{@{}c@{}}weighted\\ -F1\end{tabular} &
  acc &
  \begin{tabular}[c]{@{}c@{}}weighted\\ -F1\end{tabular} &
  acc &
  \begin{tabular}[c]{@{}c@{}}weighted\\ -F1\end{tabular} &
  acc &
  \begin{tabular}[c]{@{}c@{}}weighted\\ -F1\end{tabular} \\ \hline 
\begin{tabular}[c]{@{}c@{}}Manually Defined Prompts\\ (ChatAgri-base)\end{tabular} &
  0.798 &
  0.793 &
  0.794 &
  0.789 &
  0.866 &
  0.853 &
  \underline{0.978} &
  \underline{0.978} &
  0.863 &
  0.856 \\ \hline
\rowcolor[HTML]{FFFFFF} 
\begin{tabular}[c]{@{}c@{}}ChatGPT Triggered \\ - Prompts\end{tabular} &
  \begin{tabular}[c]{@{}c@{}}0.806\\ \textcolor[rgb]{1,0,0}{$\uparrow$ 0.8\%}\end{tabular} &
  \begin{tabular}[c]{@{}c@{}}0.803\\  \textcolor[rgb]{1,0,0}{$\uparrow$ 1.0\%}\end{tabular} &
  \begin{tabular}[c]{@{}c@{}}0.815\\  \textcolor[rgb]{1,0,0}{$\uparrow$ 2.1\%}\end{tabular} &
  \begin{tabular}[c]{@{}c@{}}0.812\\  \textcolor[rgb]{1,0,0}{$\uparrow$ 1.4\%}\end{tabular} &
  \begin{tabular}[c]{@{}c@{}}\underline{0.871}\\  \textcolor[rgb]{1,0,0}{$\uparrow$ 0.5\%}\end{tabular} &
  \begin{tabular}[c]{@{}c@{}}\underline{0.862}\\  \textcolor[rgb]{1,0,0}{$\uparrow$ 0.9\%}\end{tabular} &
  \begin{tabular}[c]{@{}c@{}}\underline{0.978}\\ \textcolor[rgb]{0.45,0.45,0.45}{= 0.0\%}\end{tabular} &
  \begin{tabular}[c]{@{}c@{}}\underline{0.978}\\ \textcolor[rgb]{0.45,0.45,0.45}{= 0.0\%}\end{tabular} &
  \begin{tabular}[c]{@{}c@{}}\underline{0.874}\\  \textcolor[rgb]{1,0,0}{$\uparrow$ 1.1\%}\end{tabular} &
  \begin{tabular}[c]{@{}c@{}}\underline{0.867}\\  \textcolor[rgb]{1,0,0}{$\uparrow$ 1.1\%}\end{tabular} \\ \hline
\rowcolor[HTML]{FFFFFF} 
\begin{tabular}[c]{@{}c@{}}Zero-Shot Similarity \\ - Prompts\end{tabular} &
  \begin{tabular}[c]{@{}c@{}}\underline{0.810}\\  \textcolor[rgb]{1,0,0}{$\uparrow$ 1.2\%}\end{tabular} &
  \begin{tabular}[c]{@{}c@{}}\underline{0.807}\\  \textcolor[rgb]{1,0,0}{$\uparrow$ 1.4\%}\end{tabular} &
  \begin{tabular}[c]{@{}c@{}}\underline{0.824}\\  \textcolor[rgb]{1,0,0}{$\uparrow$ 3.0\%}\end{tabular} &
  \begin{tabular}[c]{@{}c@{}}\underline{0.821}\\  \textcolor[rgb]{1,0,0}{$\uparrow$ 2.2\%}\end{tabular} &
  \textbf{\begin{tabular}[c]{@{}c@{}}0.874\\  \textcolor[rgb]{1,0,0}{$\uparrow$ 0.8\%}\end{tabular}} &
  \textbf{\begin{tabular}[c]{@{}c@{}}0.866\\  \textcolor[rgb]{1,0,0}{$\uparrow$ 1.3\%}\end{tabular}} &
  \begin{tabular}[c]{@{}c@{}}0.975\\  \textcolor[rgb]{0,0.75,0}{$\downarrow$ 0.3\%}\end{tabular} &
  \begin{tabular}[c]{@{}c@{}}0.975\\ \textcolor[rgb]{0,0.75,0}{$\downarrow$ 0.3\%}\end{tabular} &
  \begin{tabular}[c]{@{}c@{}}0.863\\ \textcolor[rgb]{0.35,0.35,0.35}{= 0.0\%}\end{tabular} &
  \begin{tabular}[c]{@{}c@{}}0.856\\ \textcolor[rgb]{0.35,0.35,0.35}{= 0.0\%}\end{tabular} \\ \hline
\rowcolor[HTML]{FFFFFF} 
\begin{tabular}[c]{@{}c@{}}Chain-of-Thought Triggered \\ - Prompts\end{tabular} &
  \textbf{\begin{tabular}[c]{@{}c@{}}0.816\\ \textcolor[rgb]{1,0,0}{$\uparrow$ 1.8\%}\end{tabular}} &
  \textbf{\begin{tabular}[c]{@{}c@{}}0.814\\ \textcolor[rgb]{1,0,0}{$\uparrow$ 2.1\%}\end{tabular}} &
  \textbf{\begin{tabular}[c]{@{}c@{}}0.832\\ \textcolor[rgb]{1,0,0}{$\uparrow$ 3.8\%}\end{tabular}} &
  \textbf{\begin{tabular}[c]{@{}c@{}}0.829\\ \textcolor[rgb]{1,0,0}{$\uparrow$ 3.0\%}\end{tabular}} &
  \textbf{\begin{tabular}[c]{@{}c@{}}0.874\\ \textcolor[rgb]{1,0,0}{$\uparrow$ 0.8\%}\end{tabular}} &
  \textbf{\begin{tabular}[c]{@{}c@{}}0.866\\ \textcolor[rgb]{1,0,0}{$\uparrow$ 1.3\%}\end{tabular}} &
  \textbf{\begin{tabular}[c]{@{}c@{}}0.981\\ \textcolor[rgb]{1,0,0}{$\uparrow$ 0.3\%}\end{tabular}} &
  \textbf{\begin{tabular}[c]{@{}c@{}}0.981\\ \textcolor[rgb]{1,0,0}{$\uparrow$ 0.3\%}\end{tabular}} &
  \textbf{\begin{tabular}[c]{@{}c@{}}0.889\\ \textcolor[rgb]{1,0,0}{$\uparrow$ 2.7\%}\end{tabular}} &
  \textbf{\begin{tabular}[c]{@{}c@{}}0.883\\ \textcolor[rgb]{1,0,0}{$\uparrow$ 2.7\%}\end{tabular}} \\ \hline
\end{tabular}}
\end{table*}

On the Natural-Hazards-Type disaster category classification dataset based on the transformation of Natural-Hazards-Twitter, we found that both the PLM-based method and ChatAgri performed very well, fluctuating around 94\% to 97\% of accuracy and weighted-f1,  which meets almost all the users' needs. By observing this dataset itself, we observe that most of the text in the dataset can be classified by using some fixed phrases as trigger words. For example, there is a sentence in the dataset: ``\textit{Florida governor declares state of emergency ahead of \texttt{Dorian} and warns Floridians on the East Coast}'', where the word ``\textit{\texttt{Dorian}}'' essentially belongs to the topic of a happened American hurricane disaster. As we know, a simple semantic context always can make the training and prediction of NLU tasks much simpler, so these existing SOTA models have achieved satisfactory performances.
It is worth mentioning that during the process of reorganizing the Natural-Hazards-Twitter dataset into the Natural-Hazards-Type dataset, we intuitively maintained the same quantity of test samples for each category. Therefore, the calculation results of the accuracy indicator on the Natural-Hazards-Type dataset are the same with the weighted-F1 indicator.

The above discussion fully demonstrate the superiority of ChatGPT in agricultural text classification: even though ChatGPT has not been trained on any training set, it can still outperform all kinds of SOTA methods that trained on large-scale training sets. Note that ChatAgri-base used as a comparison baseline here solely employs the manually defined prompting strategy, which is a basic and simple one. Even the simple ChatAgri can achieve impressive results, which makes us more convinced that the ChatGPT-based solution will be the future direction for the continuous research development of agricultural text classification.

\subsection{Improving ChatGPT with Advanced Prompting Strategies}
\label{subsub_aps}
In order to explore the influence of different prompt generation strategies to the final classification performance, we conducted systematic evaluations and in-depth explorations of various prompt generation strategies introduced in Section~\ref{pqc_sub} to clarify the advantages and significance of different prompt generation strategies in this section. The current date for ChatGPT interface calls is March 24, 2023. Subsequent OpenAI updates to the ChatGPT official API may influence the future function calls, leading to certain performance discrepancies.

From the first two rows of Table~\ref{table_advanceps}, it can be discovered that the ChatAgri which adopts ChatGPT Triggered-Prompts outperforms the Manually Defined Prompts strategy counterpart in most cases, indicating that ChatGPT can generate better prompts to trigger its more comprehensive knowledge for more accurate prediction. For instance, ChagAgri based on ChatGPT Triggered-Prompts improved the accuracy by average 2.1\% and 1.1\% on the PestObserver-France and Agri-News-Chinese datasets, respectively, compared to ChagAgri based on Manually Defined Prompts. This empirically demonstrates that prompt engineering for ChatGPT should be combined with ChatGPT's own understanding and feedback to achieve better classification performance.

From the third and fourth rows of Table~\ref{table_advanceps}, it can be observed that the Zero-Shot Similarity-Prompts strategy performs significantly better than the baseline prompts on the first three datasets, but its performance on the Natural-Hazards-Type and Agri-News-Chinese datasets is relatively unsatisfactory, even falling behind the basic prompts, namely Manually Defined Prompts and ChatGPT Triggered-Prompts. For example, ChatAgri based on Zero-Shot Similarity-Prompts reduced the accuracy and weighted-f1 by 0.3\% compared to ChatAgri-base based on Manually Defined Prompts on the Natural-Hazards-Type dataset.

We can also easily observe from Table~\ref{table_advanceps} that the Chain-of-Thought Prompts strategy significantly improves the overall task performance on all datasets, and its effect is better than that of ChatAgri based on Zero-Shot Similarity-Prompts. Especially on the Natural-Hazards-Type and Agri-News-Chinese datasets, Chain-of-Thought Triggered-Prompts has further improved, which is an excellent effect that Zero-Shot Similarity-Prompts cannot achieve. For example, on the Agri-News-Chinese dataset, Chain-of-Thought Triggered-Prompts simultaneously improved the accuracy and weighted-f1 by average 2.7\% compared to ChagAgri-base.

\begin{table*}[h]
\renewcommand{\arraystretch}{1.3}
\caption{Performance statistics of ChatAgri and prompt learning baselines in the zero/few-shot supervised learning. Values (\%) in \textcolor[rgb]{0,0.75,0}{\textbf{green}} represent the increased performances of ChatAgri (zero-shot) compared to the second-best results (50-shot). \label{table_fewshot}}
\centering
\scalebox{0.78}{
\begin{tabular}{cccccccccccc}
\hline
&                                                                     & \multicolumn{2}{c}{\begin{tabular}[c]{@{}c@{}}Amazon-Food\\ -Comments\end{tabular}}                                                         & \multicolumn{2}{c}{\begin{tabular}[c]{@{}c@{}}PestObserver\\ -France\end{tabular}}                                                          & \multicolumn{2}{c}{\begin{tabular}[c]{@{}c@{}}Natural-Hazards\\ -Twitter\end{tabular}}                                                      & \multicolumn{2}{c}{\begin{tabular}[c]{@{}c@{}}Natural-Hazards\\ -Type\end{tabular}}                                                         & \multicolumn{2}{c}{\begin{tabular}[c]{@{}c@{}}Agri-News\\ -Chinese\end{tabular}}                                                             \\ \cline{3-12} 
\multirow{-2}{*} {\textbf{\begin{tabular}[c]{@{}c@{}}Few-Shot \\ Learning\end{tabular}}}                         & \multirow{-2}{*}{\textbf{Methods}}                                  & acc                                 & \begin{tabular}[c]{@{}c@{}}weighted\\ -F1\end{tabular} & acc                                 & \begin{tabular}[c]{@{}c@{}}weighted\\ -F1\end{tabular} & acc                                 & \begin{tabular}[c]{@{}c@{}}weighted\\ -F1\end{tabular} & acc                                 & \begin{tabular}[c]{@{}c@{}}weighted\\ -F1\end{tabular} & acc                                 & \begin{tabular}[c]{@{}c@{}}weighted\\ -F1\end{tabular} \\ \hline
                                                                       & \begin{tabular}[c]{@{}c@{}}T5-based\\ prompt-tuning\end{tabular}    & 0.521                                                       & 0.523                                                                          & 0.474                                                       & 0.466                                                                          & 0.562                                                       & 0.545                                                                          & 0.597                                                       & 0.597                                                                          & 0.425                                                       & 0.419                                                                          \\ \cline{2-12} 
\multirow{-2}{*}{\textbf{\begin{tabular}[c]{@{}c@{}}Zero -\\ Shot\end{tabular}}}                                   & \begin{tabular}[c]{@{}c@{}}BART-based\\ prompt-tuning\end{tabular}  & 0.545                                                       & 0.539                                                                          & 0.439                                                       & 0.431                                                                          & 0.573                                                       & 0.566                                                                          & 0.639                                                       & 0.639                                                                          & 0.452                                                       & 0.447                                                                          \\ \hline
                                                                       & \begin{tabular}[c]{@{}c@{}}T5-based \\ prompt-tuning\end{tabular}   & 0.605                                                       & 0.595                                                                          & 0.585                                                       & 0.578                                                                          & 0.674                                                       & 0.651                                                                          & 0.757                                                       & 0.757                                                                          & 0.563                                                       & 0.559                                                                          \\ \cline{2-12} 
\multirow{-2}{*}{\textbf{\begin{tabular}[c]{@{}c@{}}20 -\\ Shot\end{tabular}}}                                     & \begin{tabular}[c]{@{}c@{}}BART-based \\ prompt-tuning\end{tabular} & 0.627                                                       & 0.609                                                                          & 0.563                                                       & 0.554                                                                          & 0.643                                                       & 0.626                                                                          & 0.761                                                       & 0.761                                                                          & 0.594                                                       & 0.592                                                                          \\ \hline
                                                                       & \begin{tabular}[c]{@{}c@{}}T5-based \\ prompt-tuning\end{tabular}   & 0.679                                                       & 0.674                                                                          & \underline{0.656}                                                 & \underline{0.647}                                                                    & 0.732                                                       & 0.719                                                                          & 0.831                                                       & 0.831                                                                          & \underline{0.766}                                                 & \underline{0.760}                                                                    \\ \cline{2-12} 
\multirow{-2}{*}{\textbf{\begin{tabular}[c]{@{}c@{}}50 -\\ Shot\end{tabular}}}                                     & \begin{tabular}[c]{@{}c@{}}BART-based \\ prompt-tuning\end{tabular} & \underline{0.694}                                                 & \underline{0.688}                                                                    & 0.643                                                       & 0.629                                                                          & \underline{0.758}                                                 & \underline{0.746}                                                                    & \underline{0.854}                                                 & \underline{0.854}                                                                    & 0.742                                                       & 0.738                                                                          \\ \hline
\rowcolor[HTML]{FFFFFF} 
\textbf{\begin{tabular}[c]{@{}c@{}}Zero-Shot\\ (Default)\end{tabular}} & \begin{tabular}[c]{@{}c@{}}ChatAgri-base \\ (Ours)\end{tabular}     & \textbf{\begin{tabular}[c]{@{}c@{}}0.798\\ $\uparrow$ \textcolor[rgb]{0,0.75,0}{10.5\%}\end{tabular}} & \textbf{\begin{tabular}[c]{@{}c@{}}0.793\\ $\uparrow$\textcolor[rgb]{0,0.75,0}{10.5\%}\end{tabular}}                    & \textbf{\begin{tabular}[c]{@{}c@{}}0.794\\ $\uparrow$ \textcolor[rgb]{0,0.75,0}{15.1\%}\end{tabular}} & \textbf{\begin{tabular}[c]{@{}c@{}}0.789\\ $\uparrow$\textcolor[rgb]{0,0.75,0}{16.0\%}\end{tabular}}                    & \textbf{\begin{tabular}[c]{@{}c@{}}0.866\\ $\uparrow$ \textcolor[rgb]{0,0.75,0}{10.8\%}\end{tabular}} & \textbf{\begin{tabular}[c]{@{}c@{}}0.853\\ $\uparrow$\textcolor[rgb]{0,0.75,0}{10.7\%}\end{tabular}}                    & \textbf{\begin{tabular}[c]{@{}c@{}}0.978\\ $\uparrow$ \textcolor[rgb]{0,0.75,0}{12.4\%}\end{tabular}} & \textbf{\begin{tabular}[c]{@{}c@{}}0.978\\ $\uparrow$\textcolor[rgb]{0,0.75,0}{12.4\%}\end{tabular}}                    & \textbf{\begin{tabular}[c]{@{}c@{}}0.863\\ $\uparrow$ \textcolor[rgb]{0,0.75,0}{12.1\%}\end{tabular}} & \textbf{\begin{tabular}[c]{@{}c@{}}0.856\\ $\uparrow$\textcolor[rgb]{0,0.75,0}{11.8\%}\end{tabular}}                    \\ \hline
\end{tabular}}
\end{table*}

It is worth mentioning that for the binary classification dataset Natural-Hazards-Twitter, the classification process based on the Chain-of-Thought rules only requires one comparison step, and the pivot sentence selected by this strategy is exactly the same as that used by Zero-Shot Similarity-Prompts. Therefore, the performance of the Chain-of-Thought Prompts and Zero-Shot Similarity-Prompts strategies is the same here. Moreover, due to the simple semantics of the Natural-Hazards-Type constructed by us, the prediction effect of various ChatAgri model variants is close to saturation. Therefore, the Natural-Hazards-Type dataset is not more persuasive than other datasets in terms of reference value.

In summary, Chain-of-Thought Triggered-Prompts is particularly good at handling texts with many classification categories in multi-classification tasks, which also confirms the effectiveness of the divide-and-conquer idea of splitting complex multiple classification tasks into multiple simple binary classifications for handling slightly complex classification tasks. In contrast, Zero-Shot Similarity-Prompts performs relatively poorly when there are many classification categories, and even worse than the effects of Manually Defined Prompts and ChatGPT Triggered-Prompts. We speculate that the main reason is that the selection of pivot sentences is not perfect on the one hand, and on the other hand, when ChatGPT judges the specific similarity of multiple semantically similar pivot sentences, multiple semantically similar pivot sentences can easily confuse ChatGPT, leading to its easy misjudgment of the final classification result.

\subsection{Few-shot prompt-tuning and zero-shot ChatAgri}
Although most representative text classification methods are based on supervised learning with a large volume of high-quality annotated samples. The fact is, the  annotation procedure of supervised corpora demands the expertise of domain specialists and is expensive and time-consuming, as well as a significant amount of manual efforts. Thus, in specific practical application scenarios, it is often more widespread and ubiquitous to apply data-scarce learning due to insufficient resource and scarce data. 

As numerous literature have suggested \cite{promptsurvey,ptuning,ptuningv2}, prompt-learning is particularly useful in data insufficient scenarios. It is a powerful and promising NLP technique which fully leverages the prior knowledge learned from the PLM's pre-trained stage. By using the prompting tricks, prompt-learning allows PLMs quickly adapt to various new tasks while learning on a small amount of data. Here, we delved in-depth into the characteristics, differences, and interactions between ChatGPT and prompt-learning paradigms. The evaluation statistic of these prompt learning methods was simulated based on the open-source framework \textit{OpenPrompt}. \textit{OpenPrompt} \cite{openprompt} is an advanced research toolkit developed by Tsinghua University$\footnote{OpenPrompt can be accessed at \url{https://github.com/thunlp/OpenPrompt/} [Accessed on 2023.03].}$. \textit{OpenPrompt} integrates various prompt-based learning methods, making it easy and feasible for researchers to quickly develop and deploy their prompt-tuning solutions. 

Correspondingly, we provided a detailed comparison to explore the relationships between ChatAgri and PLM-based prompt-tuning methods under few-shot and zero-shot learning settings. As shown in Table~\ref{table_fewshot}, we report the experimental results of these SOTA methods (i.e. T5-based prompt-tuning, BART-based prompt-tuning and ChatAgri) under the few-shot learning and zero-shot settings.

Specifically, from the first row of Table~\ref{table_fewshot}, it can be seen that prompt learning methods are extremely effective in zero-shot learning (i.e., without any training on any samples), far surpassing the performance of models that guess based on average probability. For instance, on the Natural-Hazards-Twitter dataset, the BART-based prompt-tuning method achieved an accuracy of 57.3\% in zero-shot learning, compared to a performance of 33.3\% based on average probability, an improvement of about 24 percentage points. Especially on the five-classification dataset, Natural-Hazards-Type, the evaluated accuracy was 63.9\%, which is much higher than the baseline accuracy of 20\% for random prediction. In addition, under the 20-shot and 50-shot few-shot settings, the improvement of these prompt learning methods is even more significant, and the specific experimental results can be found in the third and fourth rows. The above statistical results indicate that prompt learning methods are very effective in training with small amounts of data.

Most impressively, it can be obviously observed from the table that ChatAgri performs significantly better than these prompt learning methods and achieves state-of-the-art performances in most aspects, regardless of different classification category topics and counts.
The text classification performance of ChatAgri-base has surpassed these SOTA models in all test datasets with a significant improvement, demonstrating its superiority in all aspects. For example, compared with the baseline BART-based prompt-tuning that trained on 50-shot setting, ChatAgri-base yielded approximately absolute 10.5\%, 15.1\%, and 10.8\% improvements in accuracy on datasets Amazon-Food-Comment, PestObserver-France, and Natural-Hazards-Twitter, respectively. It goes without saying that even compared to prompt learning models under zero-shot learning, those better performed, which is trained on a small amount of data, are significantly inferior to the ChatGPT-based classification framework ChatAgri without any fine-tuning. In addition, better prompt engineering, ChatGPT models, and answer alignment engineering could further bring better results to the ChatAgri technology.
Overall, ChatAgri has essentially surpassed the existing state-of-the-art prompt learning paradigm in all aspects, which is also the enormous potentials brought by the ultra-large-scale models.

In conclusion, ChatAgri shows its effectiveness and superiority in data-insufficient learning scenarios, indicating that ChatGPT has strong cross-domain and generalization capabilities. This kind of generalization is one of the directions for the development of future General Purpose AI, as it can help us build more flexible and adaptable intelligent systems that can handle various tasks and scenarios.  

As we know, better performance would like to be obtained once using smoother prompts or update ChatGPT itself. As the impact of advanced prompting strategies has been investigated in Section~\ref{subsub_aps}, we then explore the potentials of upgrading the ChatAgri framework with more advanced ChatGPT, GPT-4. 


\begin{figure*}[!t]
\centering
\includegraphics[width=\linewidth]{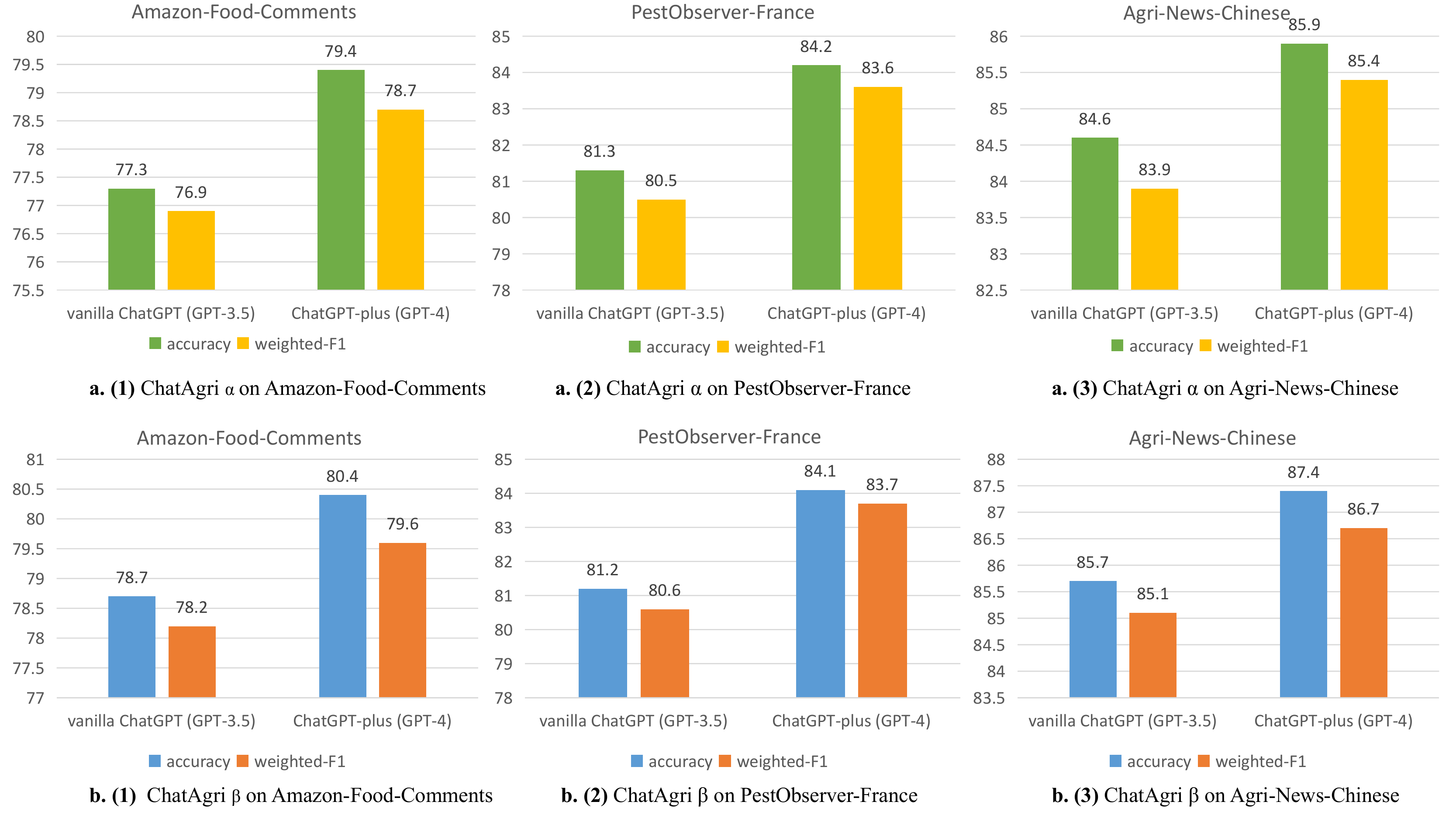}
\caption{The values shows the absolute metrics of accuracy and wighted-F1, which are reported using (\%). The first group of a.(1), a.(2) and a.(3) denotes the ChatAgri$_{\alpha}$, and the second group of b.(1), b.(2) and b.(3) denotes the ChatAgri$_{\beta}$ counterpart. Reported results were averaged over 5 runs to ensure experimental reliability and robustness.}
\label{fig_GPT4comparison}
\end{figure*}

\subsection{Potentials between ChatGPT and GPT-4}
Just as we were conducting research on vanilla ChatGPT (GPT-3.5) in March to April, 2023, OpenAI coincidentally released their latest powerful conversational system, GPT-4 \cite{GPT4}, which serves as an improved version of ChatGPT. Thus, it is necessary to conduct additional exploration experiments to evaluate the overall performance of GPT-4, the upgraded ChatGPT, in the agriculture field text classification task. 

Building on the advanced technologies learned from ChatGPT, GPT-4 has been iteratively refined to achieve unprecedented levels of authenticity, controllability, and rejection of undesirable outputs.
In terms of model parameter scale, GPT-4 is expected to have over 1 trillion parameters, a significant increase from the GPT-3.5's 175 billion parameters. This means that GPT-4 will be able to handle larger amounts of data and generate longer, more complex, coherent, accurate, diverse, and creative text. In terms of overall capability, compared to the previous version of ChatGPT, GPT-4 boasts improved performances in advanced reasoning, handling complex instructions, and demonstrating more creativity.

But GPT-4 currently has a cap of 25 messages every three hours by the latest released policy of OpenAI. It is the computation resource scarcity that caused the limited API capacity, which is far way from reaching the demand of the comprehensive experiments towards GPT-4 based ChatAgri. To overcome those pitfalls, we have taken a relatively balanced approach based on the trade-offs between experimental effectiveness and resource consumption (running time and empirical cost) in our experiments. Specifically, we made several reasonable reductions to the experiment from three perspectives: the linguistic categories, scales and their  contributions of the datasets.
The specific adjustments and arrangements for this experiment are as follows:  

\begin{itemize}
\item For dataset selection, in order to comprehensively evaluate the performance of cross-linguistic text classification tasks, we selected three datasets that represent English, Chinese, and French contexts: Amazon-Food-Comments, PestObserver-France, and Agri-News-Chinese;
\item For the specific samples to be evaluated, for each independent experiment, we randomly selected 100 samples from the original evaluation set as the evaluation subset;
\item For the selection of the baselines, we used two ChatAgri models based on manually defined prompts and prompts triggered from ChatGPT, respectively labeled as ChatAgri$_{\alpha}$ and ChatAgri$_{\beta}$;
\item To ensure the reliability and accuracy of the experimental results, we conducted 5 rounds of random screening and corresponding evaluations for each dataset, and took the average of the results from multiple rounds as the final evaluation result.
\end{itemize}

According to a series of  comparative experiments, we found that GPT-4 performs better than vanilla ChatGPT, GPT-3.5. Specifically, as illustrated in Fig.~\ref{fig_GPT4comparison}, from which we can observe that the overall performance of ChatAgri$_{\alpha}$ and ChatAgri$_{\beta}$ equipped with GPT-4 is better than the counterparts equipped with vanilla ChatGPT. For example, as shown in a. (2) of Fig.~\ref{fig_GPT4comparison}, the GPT-4 based ChatAgri$_{\alpha}$ overwhelmingly outperforms the GPT-3.5 based based ChatAgri$_{\alpha}$ by obtaining about 2.9\% and 3.1\% absolute gains of accuracy and weighted-F1 on the PestObserver-France dataset. As shown in the second group, GPT-4 also has brought a significant performance gain to ChatAgri$_{\beta}$ when compared with the vanilla ChatGPT-equipped counterpart on both the Amazon-Food-Comments and Agri-News-Chinese datasets, by achieving averaged 1.7\% absolute accuracy gains. These experiment results powerfully demonstrate that GPT-4 can further exert its potentials and gain a better semantic understanding capability in handling the agricultural text classification task.

Especially in some complex semantic scenarios, like a semantic context containing a large number of semantically similar but subtly different texts, the classification accuracy of GPT-4 is significantly higher than that of vanilla ChatGPT. These results indicate that GPT-4 has higher accuracy and robustness in handling complex semantic texts, and has a wider range of application prospects. Overall, the performance of GPT-4 is proved to be much superior and more stable than the vanilla ChatGPT. So far, we sincerely hope that in the future, OpenAI will provide greater support for the successive GPT series, including GPT-4 and even more advanced versions, so that we can fully leverage the benefits brought by advanced General Purpose AI in all aspects of future sustainable agricultural applications.  

\section{Conclusion and Outlook}
\label{conclusion}
Agricultural text classification, which serves as the basis for organizing various types of documents, is a crucial step towards managing massive and ever-increasing agricultural information. Notwithstanding, existing mainstream PLM-based classification models has faced some bottlenecks that are difficult to overcome, such as high-dependency of well-annotated corpora, cross-linguistic transferrability, and complex deployment. To our surprise, the emergence of ChatGPT has brought a turning point to this dilemma.
Despite their success, there are few to no systematic of the benefits brought by ChatGPT for the sustainable agricultural information management, especially in the research field of agricultural text classification.

In this work, we have conducted a preliminary study to explore the potentials of ChatGPT in agricultural text classification. As a result, we have proposed a novel ChatGPT-based text classification framework, namely ChatAgri. To the best of our knowledge, the proposed ChatAgri is the first study performing a qualitative analysis of text classification on ChatGPT, with a focus on the agricultural domain. 
Specifically, in our experiments, we have compared ChatAgri with various baselines relying on different learning paradigms, including traditional ML methods, such as traditional machine learning, PLM-based fine-tuning, and PLM-based prompt learning. Experiments have been performed on datasets that included various languages, such as English, French, and Chinese. Furthermore, we have developed several prompt generation strategies to better stimulate the generation potentials of ChatGPT, and to ultimately evince the effectiveness of the designed prompts. Additionally, we have further investigated the capability of the latest released ChatGPT (GPT-4) through a series of comparative experiments. Overall, the examination of the results elicited by our experiments and ablation studies have revealed the superiority of applying ChatGPT in agricultural text classification. 


It is certain that this empirical exploration has opened up new milestones for the development of various ChatGPT-based agricultural information management techniques. We look forward to proposing more applications of ChatGPT in sustainable agricultural development in the future, which will help promote the digital transformation and sustainable development of the agricultural sector. For example, ChatGPT can be used in the field of smart agriculture to help farmers better manage crops and land, thereby improving agricultural production efficiency and quality. On an overarching outlook, we hope this work has succeeded in its aim at exposing the manifold of opportunities brought by LLM for the agriculture domain, leveraging the immense knowledge currently available in databases to empower this sector with exciting opportunities to benefit from modern Artificial Intelligence advances.



\section*{Acknowledgments}

The authors would like to thank the anonymous reviewers for their helpful comments, corrections, and recommendations, which significantly improved the quality of the paper. G.Y. was supported in part by the ERC IMI (101005122), the H2020 (952172), the MRC (MC/PC/21013), the Royal Society (IEC\textbackslash NSFC\textbackslash211235), the NVIDIA Academic Hardware Grant Program, the SABER project supported by Boehringer Ingelheim Ltd, and the UKRI Future Leaders Fellowship (MR/V023799/1). J.D.S. also acknowledged support from the Spanish \emph{Centro para el Desarrollo Tecnológico Industrial} (CDTI) through the AI4ES project, and the Department of Education of the Basque Government (\emph{Eusko Jaurlaritza}) via the Consolidated Research Group MATHMODE (IT1456-22). B.Z. and W.J. were supported in part by the Natural Science Basis Research Plan in Shaanxi Province of China (Project Code: 2021JQ-061). Both the first two authors, B.Z. and W.J., made equal contributions to this work. 



\bibliographystyle{elsarticle-num} 
\bibliography{elsarticle-ref}






\end{document}